\documentclass[sigconf]{acmart}
\usepackage{algorithm}
\usepackage{algorithmic}
\usepackage{bbding}

\usepackage{multicol,multirow}
\usepackage{balance}
\usepackage{bbding}
\usepackage{comment}
\usepackage{bm}

\definecolor{dkgreen}{rgb}{0,0.6,0}
\definecolor{customgray}{rgb}{0.25,0.25,0.25}
\definecolor{customred}{rgb}{0.8,0.05,0.05}
\definecolor{customblue}{rgb}{0.05,0.05,0.8}

\DeclareMathOperator*{\argmin}{arg\,min}

\newcommand{\mE}{\mathbb{E}}
\newcommand{\mV}{\mathbb{V}}
\newcommand{\calD}{\mathcal{D}}
\newcommand{\calX}{\mathcal{X}}
\newcommand{\calA}{\mathcal{A}}

\newcommand{\mx}{\bm{x}}
\newcommand{\ma}{\bm{a}}
\newcommand{\mr}{\bm{r}}

\newcommand{\wl}{w_{1:l}(l)}

\newcommand{\trueV}{V(\pi_e)}

\newcommand{\ips}{\hat{V}_{\mathrm{IPS}} (\pi_e; \calD)}
\newcommand{\iips}{\hat{V}_{\mathrm{IIPS}} (\pi_e; \calD)}
\newcommand{\rips}{\hat{V}_{\mathrm{RIPS}} (\pi_e; \calD)}
\newcommand{\PI}{\hat{V}_{\mathrm{PI}} (\pi_e; \calD)}
\newcommand{\dr}{\hat{V}_{\mathrm{CDR}} (\pi_e; \calD, \hat{Q})}

\newcommand{\RIPS}{\hat{V}_{\mathrm{RIPS}}}
\newcommand{\DR}{\hat{V}_{\mathrm{CDR}}}

\definecolor{dkred}{rgb}{0.8,0,0}
\definecolor{dkgreen}{rgb}{0,0.4,0}
\definecolor{tickgreen}{rgb}{0,0.6,0}
\newcommand{\tick}{\textcolor{tickgreen}{\CheckmarkBold}}
\newcommand{\fail}{\textcolor{red}{\XSolidBrush}}

\newcommand{\hrefcolor}[1]{\textcolor{blue}{#1}}

\AtBeginDocument{%
  \providecommand\BibTeX{{%
    \normalfont B\kern-0.5em{\scshape i\kern-0.25em b}\kern-0.8em\TeX}}}

\copyrightyear{2022}
\acmYear{2022}
\setcopyright{acmcopyright}
\acmDOI{10.1145/3488560.3498380}

\acmConference[WSDM '22] {Proceedings of the Fifteenth ACM International Conference on Web Search and Data Mining}{February 21--25, 2022}{Tempe, AZ, USA.}
\acmBooktitle{Proceedings of the Fifteenth ACM International Conference on Web Search and Data Mining (WSDM '22), February 21--25, 2022, Tempe, AZ, USA}
\acmPrice{15.00}
\acmISBN{978-1-4503-9132-0/22/02}

\author{Haruka Kiyohara}
\authornote{This work was done at Hanjuku-Kaso Co., Ltd.}
\affiliation{Tokyo Institute of Technology}
\email{kiyohara.h.aa@m.titech.ac.jp}

\author{Yuta Saito}
\authornotemark [1]
\affiliation{Cornell University}
\email{ys552@cornell.edu}

\author{Tatsuya Matsuhiro}
\affiliation{Yahoo Japan Corporation}
\email{tmatsuhi@yahoo-corp.jp}

\author{Yusuke Narita}
\affiliation{Yale University}
\email{yusuke.narita@yale.edu}

\author{Nobuyuki Shimizu}
\affiliation{Yahoo Japan Corporation}
\email{nobushim@yahoo-corp.jp}

\author{Yasuo Yamamoto}
\affiliation{Yahoo Japan Corporation}
\email{yasyamam@yahoo-corp.jp}

\begin{document}
\fancyhead{}
\title{Doubly Robust Off-Policy Evaluation for Ranking Policies \\ under the Cascade Behavior Model}


\renewcommand{\shortauthors}{}

\begin{abstract}
In real-world recommender systems and search engines, optimizing \textit{ranking} decisions to present a ranked list of relevant items is critical.
\textit{Off-policy evaluation} (OPE) for ranking policies is thus gaining a growing interest because it enables performance estimation of new ranking policies using only logged data. 
Although OPE in contextual bandits has been studied extensively, its naive application to the ranking setting faces a critical variance issue due to the huge item space. To tackle this problem, previous studies introduce some assumptions on user behavior to make the combinatorial item space tractable. However, an unrealistic assumption may, in turn, cause serious bias. 
Therefore, appropriately controlling the bias-variance tradeoff by imposing a reasonable assumption is the key for success in OPE of ranking policies.
To achieve a well-balanced bias-variance tradeoff, we propose the \textit{Cascade Doubly Robust} estimator building on the \textit{cascade} assumption, which assumes that a user interacts with items sequentially from the top position in a ranking.
We show that the proposed estimator is unbiased in more cases compared to existing estimators that make stronger assumptions. Furthermore, compared to a previous estimator based on the same cascade assumption, the proposed estimator reduces the variance by leveraging a control variate. Comprehensive experiments on both synthetic and real-world data demonstrate that our estimator leads to more accurate OPE than existing estimators in a variety of settings.
\end{abstract}



\maketitle

\section{Introduction}
\label{sec:introduction}
Real-world recommender and search systems (e.g., e-commerce, music streaming, and news) aim to optimize \textit{ranking} decisions to increase sales or to enhance user satisfaction. For this purpose, bandit and reinforcement learning (RL) policies are often used. There is a growing interest in their offline performance evaluation. 
Offline evaluation is beneficial, as online A/B tests entail the risk of failure and may damage user satisfaction~\citep{gilotte2018offline, saito2021counterfactual, saito2021evaluating, kiyohara2021accelerating}.
Fortunately, we often have logged data for enabling an accurate offline evaluation. For example, the logs of a music streaming system records the list of recommended songs (i.e., playlists) and which songs the user listened to. This gives the system a chance to redesign the ranking policies to lead to more relevant ranked recommendations.
Exploiting logged data is, however, not as straightforward as conventional supervised machine learning because the reward is only observed for the items chosen by the past policy~\citep{gruson2019offline, saito2020open, levine2020offline}.

\textit{Off-policy evaluation} (OPE) is a technique to enable unbiased performance estimation of a counterfactual (or evaluation) policy using only logged data collected by a past (or behavior) policy. The OPE research community has made great progress in both contextual bandit and RL settings~\citep{beygelzimer2009offset, precup2000eligibility, strehl2010learning, dudik2014doubly, jiang2016doubly, thomas2016data, farajtabar2018more, su2020doubly, kallus2020optimal, saito2020doubly}. In the \textit{slate} contextual bandit setting where we present a ranked list of items, however, conventional OPE faces challenges because of a huge combinatorial item space. Indeed, naive use of OPE suffers from extreme variance in the slate contextual bandit setting~\citep{swaminathan2017off, li2018offline, mcinerney2020counterfactual}. One way to reduce the variance is introducing a reasonable assumption on user behavior to make the combinatorial item space tractable. However, unrealistically strong assumptions may cause serious bias~\citep{mcinerney2020counterfactual}. Therefore, achieving a well-balanced bias-variance tradeoff by introducing an appropriate user behavior assumption is the key for enabling accurate OPE of ranking policies.

\begin{figure*}[h]
    \centering
    \includegraphics[width=0.80\linewidth]{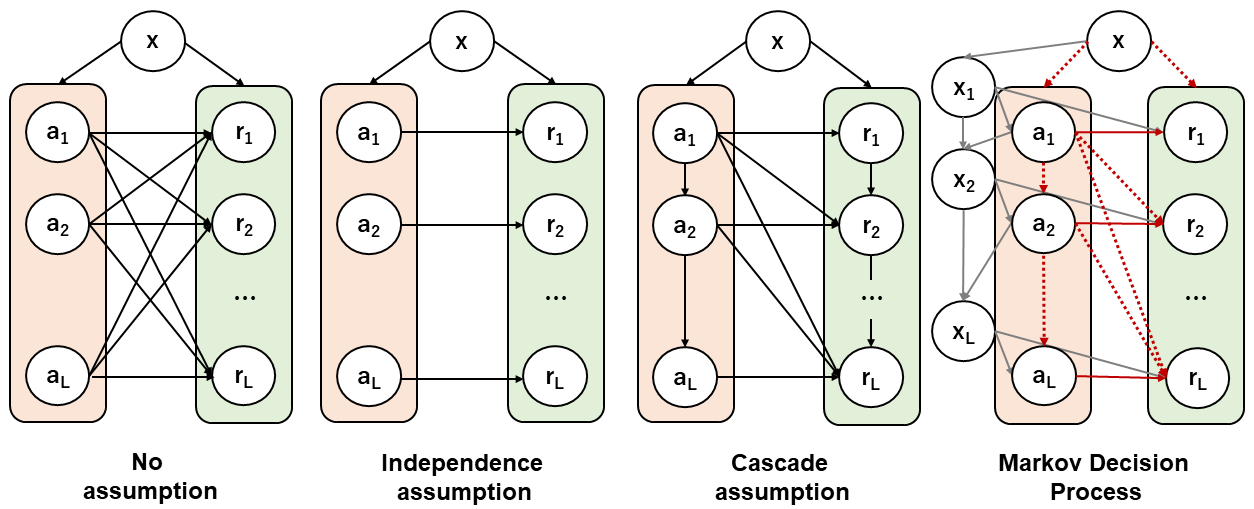}
    \caption{Comparison of user behavior assumptions and Markov Decision Process}
    \label{fig:assumptions}
    \raggedright
    \fontsize{8.5pt}{8.5pt} \selectfont \textit{Note}: We let $x$ be a user context, $a$ be an item recommended by a ranking policy, $r$ be a reward such as a click indicator, and $L$ be the slate size. Without any user behavior assumption, the reward observed at each slot is affected by all the other items in the slate. In contrast, the independence assumption assumes that a user interacts with each item independently. Therefore, the reward observed at each slot is independent of the other items. The cascade assumption assumes that a user interacts with items sequentially from the top position to the bottom. Thus, the reward observed at each slot is dependent on only the items presented at higher positions. The cascade assumption can be interpreted as a simplified version of Markov Decision Process (MDP) in RL~\citep{sutton2018reinforcement}, where we do not consider user context transitions. The red dotted arrows show the connection between the causal structural assumption of MDP and that of the cascade assumption.
\end{figure*}

\paragraph{\textbf{Previous Work}}
The prevalent method for OPE is \textit{Inverse Propensity Scoring} (IPS)~\citep{precup2000eligibility, strehl2010learning}. IPS leverages the importance sampling technique to correct bias caused by the discrepancy between behavior and evaluation policies. The benefit of IPS is that it ensures unbiasedness without any user behavior assumption. In the slate bandit setting, however, IPS can suffer from impractical variance, as the item space becomes  combinatorially large~\citep{swaminathan2017off, li2018offline, mcinerney2020counterfactual}. 
\textit{Independent IPS} (IIPS) builds on the independence assumption to address the variance issue~\citep{li2018offline}. The independence assumption assumes that a user interacts with items independently, as shown in Figure~\ref{fig:assumptions}. 
IIPS is unbiased when this assumption is satisfied and reduces the variance of IPS.
In real-world data where the assumption rarely holds, however, IIPS can yield serious bias~\citep{mcinerney2020counterfactual}.

To address the variance issue of IPS and the bias issue of IIPS, \citet{mcinerney2020counterfactual} proposed \textit{Reward interaction IPS} (RIPS), which relies on the \textit{cascade} assumption. The cascade assumption assumes that a user interacts with a list of items sequentially from the top position to the bottom~\citep{guo2009efficient}. Thus, the reward observed at each position depends only on the items at higher positions, as shown in Figure~\ref{fig:assumptions}. 
Exploiting this reasonable assumption, RIPS achieves a better bias-variance tradeoff compared to both IPS and IIPS.
Specifically, it reduces the variance of IPS and the bias of IIPS. However, RIPS can still suffer from large variance when there is a significant mismatch between the evaluation and behavior policies or the slate size is large, as we will show in the theoretical analysis.

\paragraph{\textbf{Contributions}}
To address the shortcomings of the previous estimators, we develop the \textit{Cascade Doubly Robust} (Cascade-DR) estimator for ranking policies that works under the cascade assumption.  
DR has been popular for evaluating contextual bandit and RL policies because of its desirable statistical properties~\citep{dudik2014doubly, gilotte2018offline, jiang2016doubly, thomas2016data}. This estimator improves the stability of the IPS variants, by introducing a baseline estimator as a control variate and performing propensity weighting only on its residual. 
However, deriving a DR estimator that fits the cascade assumption is non-trivial because of the sequential nature of the assumption.

In this work, we explore a way to define a DR estimator in the slate contextual bandit setting to further reduce the variance of the previous estimators.
A key trick in deriving Cascade-DR is to interpret the cascade assumption as a simplified Markov Decision Process (MDP) used in RL~\cite{sutton2018reinforcement}.
In our theoretical analysis, we prove that Cascade-DR provides unbiased OPE when the cascade assumption holds. 
This implies that the proposed estimator is unbiased in more cases compared to IIPS, as the cascade assumption is less restrictive than the independence assumption of IIPS.
Moreover, Cascade-DR reduces the variance of RIPS under a reasonable assumption about the accuracy of the baseline estimator. Thus, the proposed estimator ensures better statistical properties than the previous estimators, including IPS, IIPS, and RIPS.

To illustrate the practical benefit of our method, we conduct comprehensive experiments on both synthetic and real-world data. The results demonstrate that the proposed estimator can lead to a more accurate and stable OPE than the other estimators for ranking policies in various settings.

Our contributions are summarized as follows:
\begin{itemize}
    \item We propose Cascade-DR for OPE of ranking policies, which leverages the Markov structure of the cascade assumption.
    \item We prove the statistical advantages of the proposed estimator, including unbiasedness and variance reduction.
    \item We conduct experiments on both synthetic and real-world data. The results demonstrate that the proposed estimator outperforms the existing estimators in various situations.
\end{itemize}

\section{Preliminaries}
\label{sec:setup}
In this section, we describe the problem setting and summarize existing estimators with their statistical properties.

\subsection{Setup}
We consider a \textit{slate} contextual bandit setting. Let $\mx \in \calX \subseteq \mathbb{R}^d$ be a context vector (e.g., the user’s demographic profile) that the decision maker observes when choosing a slate action. 
Let $\calA$ be a finite set of discrete actions. Let $\ma= (a_1, a_2, \ldots, a_l, \ldots, a_L)$ be a slate action vector where  $L$ is the length of a slate (slate size).
We call a function $\pi: \calX \rightarrow \Delta (\calA^L)$ a \textit{factorizable} policy. It maps each context $\mx \in \calX$ into a distribution over the slate action, where $\pi(\ma \mid \mx) := \prod_{l=1}^L \pi(a_l \mid \mx)$ is the probability of taking slate action $\ma$ given context $\mx$. As the action selection is independent across slots, a factorizable policy can choose the same action more than twice in a slate. In contrast, we call $\pi: \calX \rightarrow \Delta (\Pi_L(\calA))$ a \textit{non-factorizable} policy, where $\Pi_L(\calA)$ is a set of $L$-permutation of $\mathcal{A}$. In this case, a policy chooses a slate action without any duplicates among slots (i.e., $\forall 1 \leq k < l \leq L, a_k \neq a_l$).
We use $\mr = (r_1, r_2, \ldots, r_l, \ldots, r_L)$ to denote a reward vector, where $r_l$ is a random variable representing the \textbf{slot}-level reward observed at slot $l$ (e.g., whether the item recommended at slot $l$ results in a click). 
Following \citet{mcinerney2020counterfactual}, we assume that the \textbf{slate}-level reward $r^{\ast}$ is a weighted sum of the slot-level rewards, i.e., 
$$r^{\ast} = \sum_{l=1}^L \alpha_l r_l, $$ 
where $\alpha_l$ denotes a non-negative weight for slot $l$. 
Our definition of the slate-level reward captures a wide variety of information retrieval metrics.
For example, when $\alpha_l := 1/\log_2 (l+1)$, $r^{\ast}$ is called the discounted cumulative gain (DCG)~\citep{jarvelin2002cumulated}. Regarding the action and reward vectors ($\ma, \mr$), we use the following notations.
\begin{itemize}
    \item partial set of slate actions: $\ma_{l_1:l_2}:=(a_{l_1}, a_{l_1+1}, \ldots, a_{l_2-1}, a_{l_2})$
    \item partial set of slate rewards: $\mr_{l_1:l_2}:=(r_{l_1}, r_{l_1+1}, \ldots, r_{l_2-1}, r_{l_2})$
\end{itemize}

Let $\calD := \{(\mx^{(i)},\ma^{(i)},\mr^{(i)})\}_{i=1}^n$ be logged bandit data with $n$ independent observations.
$\ma^{(i)}$ is a vector of discrete variables indicating which slate action is chosen for individual $i$. 
$\mx^{(i)}$ and $\mr^{(i)}$ denote context and reward vectors observed for $i$.
We assume that a logged bandit dataset is generated by a \textit{behavior policy} $\pi_b$:
\begin{align*}
  & \{(\mx^{(i)},\ma^{(i)},\mr^{(i)})\}_{i=1}^n \\
  & \quad \sim \prod_{i=1}^n p(\mx^{(i)}) \pi_b (\ma^{(i)} \mid \mx^{(i)}) p(\mr^{(i)} \mid \mx^{(i)}, \ma^{(i)}).
\end{align*}
Note that, throughout the paper, we assume that there is no unobserved confounder, the logged dataset has full support over slate actions, and slot-level rewards ($r_1, \ldots, r_L$) are all observable.

For a function $f(\mx,\ma,\mr)$, we use
\begin{align*}
  \mE_{n} [f(\mx, \ma, \mr)] := n^{-1} \sum_{(\mx^{(i)}, \ma^{(i)}, \mr^{(i)}) \in \calD} f(\mx^{(i)}, \ma^{(i)}, \mr^{(i)})
\end{align*}
to denote its empirical expectation over observations in $\calD$. We also let $\mE_{\calD}[\cdot] := \mE_{\calD \sim \prod_{i=1}^n p(\mx^{(i)}) \pi_b(\ma^{(i)} \mid \mx^{(i)}) p(\mr^{(i)} \mid \mx^{(i)}, \ma^{(i)})}[\cdot]$.
Finally, we use $q(\mx,\ma):=\mE_{\mr \sim p(\mr|\mx,\ma)} [r^{\ast} \mid \mx, \ma ]$ to denote the \textbf{slate}-level mean reward function and $q_{l}(\mx,\ma):=\mE_{\mr \sim p(\mr|\mx,\ma)} [r_l \mid \mx, \ma ]$ to denote the \textbf{slot}-level mean reward function.

\subsection{Estimation Target}
We are interested in using logged bandit data to estimate the following \textit{policy value} of any given \textit{evaluation policy} $\pi_e$ which might be different from $\pi_b$: 
\begin{align*}
    \trueV 
    := \mE_{(\mx,\ma,\mr) \sim p(\mx) \pi_e (\ma \mid \mx) p(\mr \mid \mx, \ma)} [r^{\ast}].
\end{align*}
Estimating $\trueV$ before implementing $\pi_e$ in an online environment is valuable because $\pi_e$ may perform poorly and damage user satisfaction~~\citep{gilotte2018offline, gruson2019offline, saito2020open, saito2021counterfactual, kiyohara2021accelerating}.

\subsection{Existing Estimators} \label{subsec:estimators}
Here, we review existing estimators in the slate contextual bandit setting and state their limitations.\footnote{Note that the pseudoinverse (PI) estimator has also been proposed for OPE in the slate contextual bandit setting~\citep{swaminathan2017off, vlassis2021off}. However, it assumes that only slate-level reward $r^{\ast}$ is observable, which is different from our setting. \citet{mcinerney2020counterfactual} empirically confirms that PI does not work in our setting. Thus, we do not consider PI.}

\paragraph{Inverse Propensity Scoring} 
IPS~\citep{precup2000eligibility, strehl2010learning} uses the importance sampling technique to estimate the policy value as follows:
\begin{align*}
    \ips 
    &:= \mE_n [w(\mx, \ma) r^{\ast} ]
    = \mE_{n} \left[w(\mx, \ma) \sum_{l=1}^L \alpha_l r_l \right], 
\end{align*}
where $w(\mx, \ma) := \pi_e(\ma \mid \mx) / \pi_b(\ma \mid \mx)$ is the importance weight. When the policies are factorizable, IPS becomes 
\begin{align*}
    \ips 
    & = 
    \mE_n \left[ \left(\prod_{l=1}^L \frac{\pi_e(a_l \mid \mx)}{\pi_b(a_l \mid \mx)}\right) \sum_{l=1}^L \alpha_l r_l \right].
\end{align*}
IPS is unbiased and consistent without making any assumption on user behavior.
However, it has an extremely large variance when the action space ($|\calA^L|$ or $|\Pi_L(\calA)|$) is large~\citep{swaminathan2017off, li2018offline, mcinerney2020counterfactual, saito2020unbiased}.

\paragraph{Independent Inverse Propensity Scoring} 
IIPS is based on the independence assumption, i.e., a user interacts with items (actions) independently across slots~\citep{li2018offline}.\footnote{In~\citep{li2018offline}, the independence assumption is called the item-position model.} This implies that the reward observed at each slot depends only on the corresponding item and its position, ignoring the possible interactions among the items presented in the same slate. This assumption allows us to simplify the slot-level mean reward function as follows.
\begin{align*}
q_{l} (\mx, \ma) & = \mE_{r_l \sim p(r_l \mid \mx, a_l)} [r_l \mid \mx, a_l ].
\end{align*}

Based on this assumption, IIPS estimates the policy value as
\begin{align*}
    \iips 
    &:= \mE_n \left[\sum_{l=1}^L w_{l} (\mx, a_l) \alpha_l r_l \right] \\
    & = \mE_n \left[\sum_{l=1}^L \left( \frac{ \pi_e(a_l \mid \mx)}{\pi_b(a_l \mid \mx)} \right) \alpha_l r_l \right], 
\end{align*}
where $w_{l} (\mx, a_l) := \pi_e(a_l \mid \mx) / \pi_b(a_l \mid \mx)$ is the importance weight at slot $l$.\footnote{$\pi(a \mid \mx) := \sum_{\ma^{\prime}} \pi(\ma^{\prime} \mid \mx)\mathbb{I} \{ a^{\prime}_l=a \}$, where $\mathbb{I} \{ \cdot \}$ is the indicator function.}

IIPS is unbiased under the independence assumption and greatly reduces the variance of IPS.
However, when this assumption is violated, IIPS yields severe bias even if we increase the data~\citep{mcinerney2020counterfactual}.

\paragraph{Reward interaction Inverse Propensity Scoring} 
RIPS~\citep{mcinerney2020counterfactual} assumes that a user examines items in a ranking one-by-one from the top position~\citep{guo2009efficient}. 
Under this cascade assumption, the reward at each position ($r_l$) is dependent only on items and rewards at higher positions ($\ma_{1:l}, \mr_{1:l-1}$), but is independent of the others ($\ma_{l+1:L}, \mr_{l+1:L}$). Thus, the slot-level mean reward function can be expressed as follows.
\begin{align*}
    q_{l} (\mx, \ma) 
    & = \mE_{r_{l} \sim p(r_{l} \mid \mx,\ma_{1:l}, \mr_{1:l-1})} [r_{l} \mid \mx, \ma_{1:l}, \mr_{1:l-1} ].
\end{align*}

Then, RIPS estimates the policy value as
\begin{align*}
    \rips & := \mE_n \left[\sum_{l=1}^L w_{1:l} (\mx,\ma_{1:l}) \alpha_l r_l \right] \\
    &= \mE_n \left[ \sum_{l=1}^L \left(\prod_{l'=1}^l \frac{\pi_e(a_{l'} \mid \mx, \ma_{1:l'-1})}{\pi_b(a_{l'} \mid \mx, \ma_{1:l'-1})}\right) \alpha_l r_l \right],
\end{align*}
where $w_{1:l} (\mx,\ma_{1:l}) := \pi_e(\ma_{1:l} \mid \mx) / \pi_b(\ma_{1:l} \mid \mx)$ is the importance weight at slot $l$ given previous actions ($\ma_{1:l-1}$).\footnote{$\pi(\ma_{1:l} \mid x) :=  \prod_{l'=1}^l \pi(a_{l'} \mid \mx, \ma_{1:l'-1})$.} 

RIPS is unbiased under the cascade assumption, which includes the independence assumption of IIPS as a special case.
Moreover, RIPS has a smaller variance than IPS.
However, this estimator may still be subject to the severe variance issue, when behavior and evaluation policies greatly diverge or the slate size is large.

We now formally investigate the variance of RIPS based on its recursive expression. This kind of variance analysis of RIPS is not given in the previous work~\citep{mcinerney2020counterfactual}.
For the analysis, let us define $\RIPS^{L+1-l} := \sum_{l'=l}^L ( \prod_{l''=l}^{l'} w_{1:l''}(l'') ) \alpha_{l'} r_{l'}$ where there are $L+1-l$ remaining slots. Note that $\RIPS^{0}:=0$ and $w_{1:l}(l) := \pi_e(a_l \mid \mx, \ma_{1:l-1}) / \pi_b(a_l \mid \mx, \ma_{1:l-1})$.
Then, for $l = 1, \ldots, L$, we have
\begin{equation}
    \RIPS^{L+1-l} := w_{1:l}(l) \left( \alpha_l r_l + \RIPS^{L-l} \right), \label{eq:recursive_rips}
\end{equation}
which corresponds to Eq.(4) of~\citep{mcinerney2020counterfactual}. $\RIPS^{L+1-l}$ estimates $V^{L+1-l} := \mE_l[ \sum_{l'=l}^L \alpha_{l'} r_{l'} ]$.\footnote{We use $\mE_l [ \cdot ] := \mE_{(\ma_{l:L}, \mr_{l:L}) \sim \pi_e(\ma_{l:L} | \mx, \ma_{1:l-1}) p(\mr_{l:L} | \mx, \ma_{1:l}, \mr_{1:l-1})} [ \cdot | \mx, \ma_{1:l-1}, \mr_{1:l-1}]$ and $\mV_l(\cdot)$ as the corresponding conditional variance.} In addition, $\RIPS^{L}$ is equivalent to $\RIPS$. 
Based on this recursive expression, we derive the variance of RIPS.

\begin{proposition} \label{prop:rips} (Variance of RIPS)
Under the cascade assumption, the (conditional) variance of RIPS is recursively given by
\begin{align*}
    & \mV_l \left( \RIPS^{L+1-l} \right)  \\
    &= \mE_l \left[ \wl^2 \mV_{l+1} \left( \RIPS^{L-l} \right) \right] 
     + \alpha_l^2 \mE_l[\wl^2 \mV_{r_l}(r_l)] \\
    & \quad + 2 \alpha_l \mE_l \Bigl[ \wl^2 \left( r_l - q_l(\mx, \ma_{1:l}) \right)  \left. \left( \RIPS^{L-l} - V^{L-l} \right) \right] \\
    & \quad + \textcolor{red}{\mV_l \left( \wl Q_l \right)}
\end{align*}
where $Q_l = Q_l(\mx, \ma_{1:l}) := \alpha_l q_l(\mx, \ma_{1:l}) + V^{L-l}$. $\mV_{r_l}(\cdot)$ denotes conditional variance of $r_l$ given $(\mx, \ma_{1:l}, \mr_{1:l-1})$. 
We provide all the proofs in the main text in Appendix~\ref{app:proof}.  
\end{proposition}

\noindent In Proposition~\ref{prop:rips}, $\mV_l \left( \wl Q_l \right)$ is the most important term as its scale is large and is recursively weighted. This becomes problematic when the slate size is large and there is a weak overlap between behavior and evaluation policies.

\section{Our Approach} \label{sec:method}
In this section, we propose a new estimator for evaluating counterfactual ranking policies.

\subsection{Cascade Doubly Robust Estimator} 
A key insight in deriving the proposed estimator is that the causal structure of the cascade assumption is similar to that of MDP in RL~\citep{sutton2018reinforcement}. Specifically, we can interpret the cascade assumption as a simplified MDP where we do not explicitly consider the user context transition.

The recursive expression of RIPS in Eq.~\eqref{eq:recursive_rips} leads to the recursive expression of Cascade-DR as follows.
\begin{align}
    & \DR^{L+1-l} := w_{1:l}(l) \left( \alpha_l r_l + \DR^{L-l} - \hat{Q}_l \right) + \mE_{a^{\prime}_l} \left[ \hat{Q}_{l} \right],  \label{eq:recursive_dr}
\end{align}
where we introduce $\hat{Q}_l$ (an estimator of $Q_l$) as a control variate. CDR stands for Cascade-DR and $\mE_{a^{\prime}_l} [\cdot] := \mE_{a_{l}^{\prime} \sim \pi_e(a_l^{\prime} \mid \mx, \ma_{1:l-1})} [\cdot]$.
Intuitively, adding the control variate makes the scale of the first term of Eq.~\eqref{eq:recursive_dr} much smaller than Eq.~\eqref{eq:recursive_rips}, if $\hat{Q}_l$ is accurate. Thus, we expect a variance reduction, which we formally discuss in Section~\ref{sec:theory}.

We now derive Cascade-DR by solving the recursive expression:
\begin{align*}
    \dr 
     := \mE_n \left[\sum_{l=1}^L \left(w_{1:l} \left( \alpha_l r_l  - \hat{Q}_{l} \right)  + w_{1:l-1} \mE_{a^{\prime}_l} \left[ \hat{Q}_{l} \right] \right) \right],
\end{align*}
where $w_{1:l} := w_{1:l}(\mx, \ma_{1:l})$.

To use Cascade-DR, we have to derive $\hat{Q}_l$ from logged bandit data. We can take advantage of the recursive structure of the reward and obtain $\hat{Q}_l$ as follows.
\begin{align*}
    \hat{Q}_l  \leftarrow \argmin_{Q_l} \mE_n \biggl[ & w_{1:l} \Bigl( Q_l(\mx, \ma_{1:l}) \\
    &\left.\left. - \left( \alpha_l r_l + \mE_{a^{\prime}_{l+1}} \left[ \hat{Q}_{l+1}(\mx, \ma_{1:l+1}^{\prime}) \right] \right) \right)^2 \right],
\end{align*}
where we set $\hat{Q}_{L+1} = 0$. $w_{1:l}$ is used to mitigate the bias from the baseline estimation.

\subsection{Theoretical Analysis} \label{sec:theory}
Cascade-DR has some desirable statistical properties.
We first show that Cascade-DR is unbiased when the cascade assumption holds.

\begin{proposition} \label{prop:unbiased} (Unbiasedness of Cascade-DR) 
Under the cascade assumption, Cascade-DR is statistically unbiased, i.e., for any given $\pi_e$ and $\hat{Q}$, we have $\mE_{\calD} \left[ \dr \right] = \trueV$.
\end{proposition}

\noindent Unbiasedness under the cascade assumption is a desirable statistical property, as the cascade assumption is more realistic compared to the independence assumption of IIPS.
Specifically, unbiasedness under the cascade assumption is a sufficient condition for unbiasedness under the independence assumption.
Therefore, Proposition~\ref{prop:unbiased} ensures that the proposed estimator is unbiased in more cases compared to IIPS.

We also show that Cascade-DR has a preferable variance.

\begin{theorem} \label{thrm:dr} (Variance of Cascade-DR) 
Under the cascade assumption, the (conditional) variance of Cascade-DR is recursively given by
\begin{align*}
    & \mV_l \left( \DR^{L+1-l} \right) \\
    &= \mE_l \left[ \wl^2 \mV_{l+1} \left( \DR^{L-l} \right) \right] 
    + \alpha_l^2 \mE_l[\wl^2 \mV_{r_l}(r_l)] \\
    & \quad + 2 \alpha_l  \mE_l \Bigl[ \wl^2 \left( r_l - 
    q_l(\mx, \ma_{1:l}) \right)  \left. \left( \DR^{L-l} - V^{L-l} \right) \right] \\
    & \quad + \textcolor{red}{\mV_l \left( \wl \Delta_l \right)}
\end{align*}
where $\Delta_l := Q_l - \hat{Q}_l$ is the estimation error of $\hat{Q}$. Note that the variance of Cascade-DR is equivalent to that of RIPS when $\hat{Q}_l=0, \, \forall l = 1, \ldots, L$.
\end{theorem}

\noindent Theorem~\ref{thrm:dr} implies that Cascade-DR improves the last term from $\mV_l (\wl Q_l)$ to $\mV_l (\wl \Delta_l)$ compared to RIPS (see Proposition~\ref{prop:rips}). Specifically, if $|Q_l -\hat{Q}_l| < Q_l, \forall l = 1, \ldots, L$, Cascade-DR achieves a smaller variance than RIPS. The condition for variance reduction (i.e., $|Q_l -\hat{Q}_l| < Q_l$) can also be represented as $0 < \hat{Q}_l < 2Q_l$, suggesting that the estimation error of $\hat{Q}_l$ should be within $\pm 100\%$. Therefore, we argue that Cascade-DR is likely to achieve a smaller variance than RIPS.

\section{Experiments: Synthetic Data}
In this section, we empirically compare the proposed estimator with the existing estimators using synthetic data.\footnote{Cascade-DR is implemented in OBP as \textbf{obp.ope.SlateCascadeDoublyRobust}. Our experimental code is available at \href{https://github.com/aiueola/wsdm2022-cascade-dr}{\texorpdfstring{\hrefcolor{https://github.com/aiueola/wsdm2022-cascade-dr}}}.}

\begin{table*}[htb]
\caption{Relationship among user behavior assumptions, OPE estimators, and reward structures}
\scalebox{0.95}{
\begin{tabular}{cc|ccc}
\toprule & \multicolumn{2}{c}{} & \textbf{Reward Structures} \\ \cmidrule{3-5}
\textbf{User Behavior Assumptions} & \textbf{OPE Estimators} & standard & cascade & independence \\ \midrule \midrule
none & IPS & \tick & \tick & \tick \\
cascade & RIPS, Cascade-DR (ours) & \fail & \tick & \tick \\
independence & IIPS & \fail & \fail & \tick \\
\bottomrule
\end{tabular} \label{tab:user_behavior}
}
\vskip 0.1in
\raggedright
\fontsize{8.5pt}{8.5pt}\selectfont \textit{Note}:
\tick means that the user behavior assumption captures the reward structure of the data generating process. When the user behavior assumption captures the reward structure, the corresponding OPE estimators are unbiased.
\end{table*}

\begin{table*}[htb]
\caption{Choices of experimental configurations in the synthetic experiment} \label{tab:configurations}
\begin{tabular}{c|c|c}
\toprule
\textbf{Configurations} & \textbf{Choices} & \textbf{Notations} \\\midrule \midrule
data size & $n \in \{250, 500, 1000, 2000, 4000\}$ & \multirow{4}{*}{$\Phi$ (to define $\calD$)}\\
slate size & $L \in \{3, 4, 5, 6, 7\}$ & \\
\multirow{2}{*}{reward\_structure} & $\text{reward\_structure} \, (F) \in \{ \text{standard}, \text{cascade}, \text{independence} \}$ & \\
& $\text{interaction\_function} \, (G) \in \{ \text{additive}, \text{decay} \}$ & \\ \midrule
policy similarity & $\lambda \in \{ -0.8, -0.6, -0.4, -0.2, 0.0, 0.2, 0.4, 0.6, 0.8 \}$ & $\Lambda$ (to define $\pi_e$) \\
\bottomrule
\end{tabular}
\end{table*}

\subsection{Setup} \label{sec:synthetic_setup}
Our synthetic experiment is based on \textit{OpenBanditPipeline} (OBP)\footnote{\href{https://github.com/st-tech/zr-obp}{\texorpdfstring{\hrefcolor{https://github.com/st-tech/zr-obp}}}} provided by \citet{saito2020open}. OBP is an open-source toolkit for OPE, which includes synthetic data generation modules for the slate contextual bandit setting. We synthesize datasets based on various user behaviors to evaluate how the estimators perform under different assumptions. We also vary the slate size and evaluation policy to evaluate the estimators' robustness to possible configuration changes~\citep{saito2021evaluating}. Below, we describe our experimental setup in detail.

\paragraph{\textbf{Basic synthetic setting}}
To generate synthetic data, we need to sample $(\mx, \ma, \mr)$. First, we randomly generate five-dimensional contexts ($d=5$), independently and normally distributed with zero mean. Then, we set $|\calA| = 5$ and $\alpha_l = 1, \forall l = 1, \ldots, L$. Finally, we sample slot-level reward $r_l$ from the Bernoulli distribution as $r_l \sim Bern(q_l(\mx, \ma))$. Note that $q_l(\mx, \ma)$ is the slot-level mean reward function. Below, we describe how to define $q_l(\mx, \ma) (= \mE[r_l \mid \mx, \ma])$ to simulate different user behavior assumptions. 

\paragraph{\textbf{Reward structures}} 
When generating synthetic data, we use three different reward structures, \textit{standard}, \textit{cascade}, and \textit{independence}. These reward structures correspond to the user behavior assumptions of different estimators as shown in Table~\ref{tab:user_behavior}.

First, we define the following slot-level base reward function. 
\begin{align*}
    \tilde{q}_l(\mx, a_l) :=\theta_{a_l}^{\top} \mx + b_{a_l},
\end{align*}
where $\theta_{a_l}$ is a parameter vector sampled from the standard normal distribution. $b_{a_l}$ is a bias term that corresponds to action $a_l$.

Next, we define the general form of the synthetic slot-level mean reward function as follows.
\begin{align}
    q_l(\mx, \ma) := \sigma \left( \tilde{q}_l(\mx, a_l) + F (\mx, \ma)  \right), \label{eq:q_func}
\end{align}
where $\sigma(z):= 1/(1+\exp(-z))$ is the sigmoid function.
Note here that $\tilde{q}_l(\mx, a_l)$ depends only on the action presented at slot $l$, while $F (\mx, \ma)$ depends on the whole slate $\ma$. 
We use three different $F(\cdot, \cdot)$ to switch the reward structures as follows.
\begin{align*}
    F (\mx, \ma) = 
    \begin{cases}
        \sum_{k \neq l} G(k, l) & \text{(standard)} \\
        \sum_{k < l} G(k, l) & \text{(cascade)} \\
        0 & \text{(independence)} \\
    \end{cases},
\end{align*}
where $G(k, l)$ defines the interaction between actions presented at slots $k$ and $l$. 
The simplest \textit{independence} reward structure assumes that the slot-level reward depends only on the corresponding slot. Therefore, there is no interactions among actions in the slate, and $q_l(\mx, \ma)$ is derived directly from $\tilde{q}_l(\mx, a_l)$. In contrast, the \textit{standard} and \textit{cascade} reward structures assume that there are some interactions among actions presented in the same slate. Specifically, the \textit{standard} reward structure assumes that $q_l(\mx, \ma)$ depends on all the other actions in the slate (i.e., $\forall a_k (k \neq l)$). On the other hand, when the reward structure is \textit{cascade}, $q_l(\mx, \ma)$ contains interactions from only the actions presented at higher slots (i.e., $\forall k < l$). This is because the \textit{cascade} assumption indicates that a user examines the slots sequentially from top to bottom.

Next, $G(k, l)$ defines the level of interaction between two slots. We use two different functions as $G(\cdot, \cdot)$. The first one is \textit{additive}. It assumes that the slot-level reward is additively affected by the co-occurrence of the two actions as 
$
    G(k, l) = \mathbb{W}(a_k, a_l)
$.
Here, $\mathbb{W}$ is $|\calA| \times |\calA|$ symmetric matrix which defines how an action affects the reward of the other actions in the same slate. 
Another variant of $G(k, l)$ is \textit{decay}. It assumes that the slot-level reward is affected by the expected rewards of neighboring actions. For example, an item may be less likely to be clicked if the neighboring items have high click probabilities. We can simulate this situation by defining $G(\cdot, \cdot)$ as
$
    G(k, l) = - h(k, l) \tilde{q}_k(\mx, a_k)
$,
where $h(k, l)$ is a function to define the decay effect. Specifically, we use $h(k, l) = (|k-l|+1)^{-1}$.

\paragraph{\textbf{Behavior and evaluation policies}}
We use factorizable contextual policies as behavior and evaluation policies. Thus, these policies can choose the same item more than twice in a slate. 

We first define a behavior policy as follows.
\begin{align}
    \pi_b(\mx, \ma) = \prod_{l=1}^L \pi_b (\mx, a_l) = \prod_{l=1}^L \operatorname{softmax} \left( f_b(\mx, a_l) \right), \label{eq:pi_b}
\end{align}
where $f_b(\mx, a_l) = \theta_{a_l}^{\top} \mx + b_{a_l}$. We sample both $\theta_{a_l}$ and $b_{a_l}$ from the standard uniform distribution.

We then generate an evaluation policy based on the (pre-defined) behavior policy as follows. 
\begin{align}
    \pi_e(\mx, \ma) = \prod_{l=1}^L \operatorname{softmax} \left(\lambda \cdot f_b(\mx, a_l) + (1 - |\lambda|) \right), \label{eq:pi_e}
\end{align}
where $\lambda \in [-1.0, 1.0)$ is an experimental hyperparameter to control the similarity between $\pi_b$ and $\pi_e$. A positive value of $\lambda$ leads to an evaluation policy that is similar to the behavior policy. On the other hand, a negative $\lambda$ leads to an evaluation policy that is dissimilar to the behavior policy.
When $\lambda = 0.0$, the evaluation policy is identical to the uniform random policy.

\paragraph{\textbf{Compared estimators}}
We compare IPS, IIPS, RIPS, and Cascade-DR (our proposal). We use the true action choice probabilities of the behavior policy $\pi_b$ to define the above estimators. To obtain $\hat{Q}_l$ of Cascade-DR, we use a Decision Tree\footnote{sklearn.tree.DecisionTreeRegressor(max\_depth=3, random\_state=12345)} implemented in \textit{scikit-learn}\footnote{\href{https://github.com/scikit-learn/scikit-learn}{\texorpdfstring{\hrefcolor{https://github.com/scikit-learn/scikit-learn}}}}.

\subsection{Study Design}
We conduct three experiments to evaluate how the performance of the estimators changes with different \textbf{(i) data size ($n$)}, \textbf{(ii) slate size ($L$)}, and \textbf{(iii) policy similarity ($\lambda$)}.

Specifically, to investigate the effect of \textbf{data size ($n$)}, we fix slate size $L=5$ and vary $n \in \{250, 500, 1000, 2000, 4000\}$.
Then, to see how the estimators perform with different \textbf{slate size ($L$)}, we fix data size $n=1000$ and vary slate size $L \in \{3, \ldots, 7\}$.
Finally, we fix data size $n=1000$ and vary the \textbf{policy similarity ($\lambda$)} as $\lambda \in \{-0.8, -0.6, \ldots, 0.8\}$.
During the experiments, we randomly sample the other configurations in Table~\ref{tab:configurations} to identify an estimator robust to possible configuration changes as done in~\citet{saito2021evaluating}.

For experiments \textbf{(i)-(iii)}, we apply Algorithm~\ref{algo:evaluation} to each estimator and obtain the set of squared errors (SE) $\mathcal{Z}$ with random seeds $s \in \{0, \ldots, 9999\} (= \mathcal{S})$.
Specifically, for every random seed $s$, we first sample experimental configurations $\phi \in \Phi$ and $\lambda \in \Lambda$.
$\Phi$ and $\Lambda$ are defined in Table~\ref{tab:configurations}.
Then, we generate synthetic data $\calD$ and evaluation policy $\pi_e$ based on sampled $\phi$ and $\lambda$. Finally, we calculate SE as $\mathrm{SE} (\hat{V}) = (V(\pi_e) - \hat{V}(\pi_e; {\calD}))^2$.
After obtaining $\mathcal{Z}$ from Algorithm~\ref{algo:evaluation}, we evaluate the performance of estimators by \textit{mean-squared-error} (MSE), $\mathrm{MSE}(\hat{V}) = |\mathcal{Z}|^{-1} \sum_{z \in \mathcal{Z}} z $. A lower value of MSE indicates that the estimator is more accurate.

\begin{algorithm}[tb]
\caption{Experimental Procedure with Synthetic Data} \label{algo:evaluation} 
  \begin{algorithmic}[1]
    \REQUIRE SyntheticDataGenerator, EvaluationPolicyFunction, estimator to be evaluated $\hat{V}$, set of data configurations $\Phi$ (i.e., set of data size $n$, slate size $L$, reward structure $F$, and interaction function $G$), set of policy similarity $\Lambda$, set of random seeds $\mathcal{S}$
    \ENSURE set of squared errors (SE), $\mathcal{Z}$
    \STATE $\mathcal{Z} \leftarrow \emptyset$ 
    \FOR{$s \in \mathcal{S}$}
      \STATE $\phi, \lambda \leftarrow \mathrm{Unif}(\Phi, \Lambda; s)$
      \STATE $\pi_b, \calD \leftarrow \mathrm{SyntheticDataGenerator}(\phi, s)$ (Eq.(\ref{eq:q_func}), Eq.(\ref{eq:pi_b}), etc.)
      \STATE $\pi_e \leftarrow \mathrm{EvaluationPolicyFunction}(\pi_b, \lambda)$ (Eq.(\ref{eq:pi_e}))
      \STATE $\mathrm{SE} (\hat{V}) = (V(\pi_e) - \hat{V}(\pi_e; \calD))^2$
      \STATE $\mathcal{Z} \leftarrow \mathcal{Z} \cup \{\mathrm{SE} (\hat{V})\}$
    \ENDFOR
  \end{algorithmic}
\end{algorithm}

\begin{figure*}[!htb]
\scalebox{0.95}{
\begin{tabular}{ccc}
\toprule
\multicolumn{3}{c}{reward structure} \\ \midrule
\textbf{standard} & \textbf{cascade} & \textbf{independence} \\ \midrule \midrule
\begin{minipage}{0.33\hsize}
    \begin{center}
        \includegraphics[clip, width=5.5cm]{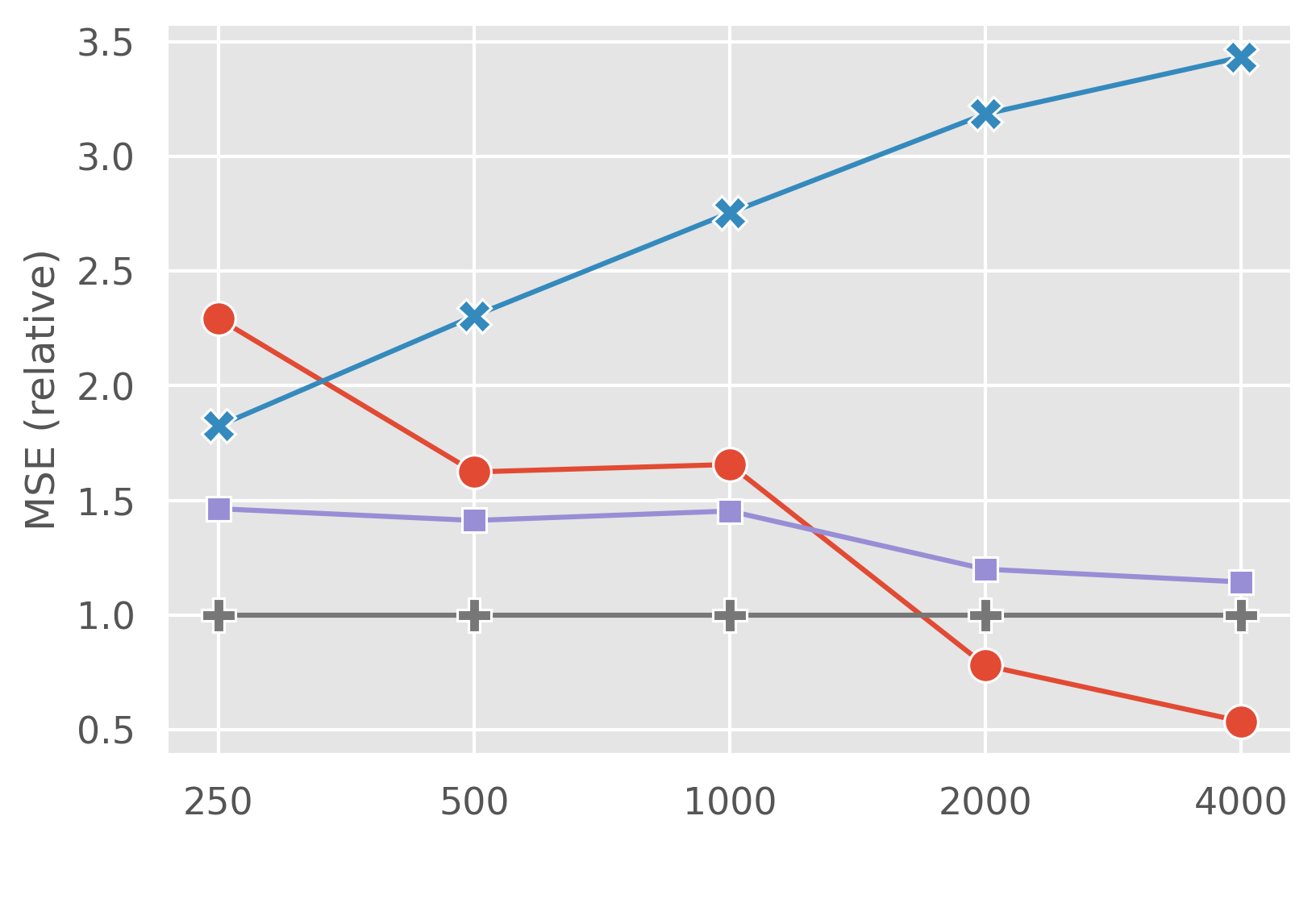}
    \end{center}
\end{minipage}
&
\begin{minipage}{0.33\hsize}
    \begin{center}
        \includegraphics[clip, width=5.5cm]{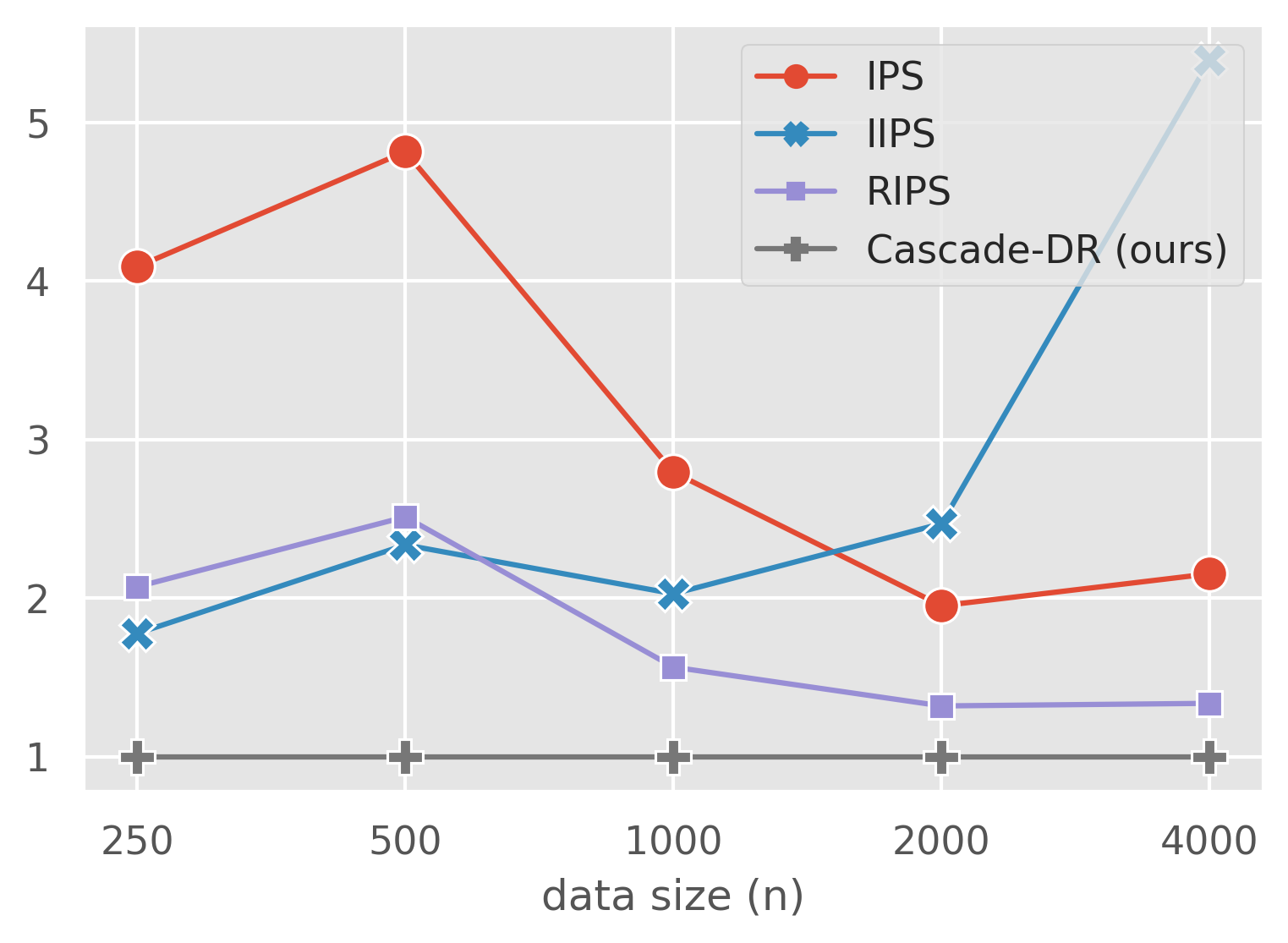}
    \end{center}
\end{minipage}
&
\begin{minipage}{0.33\hsize}
    \begin{center}
        \includegraphics[clip, width=5.5cm]{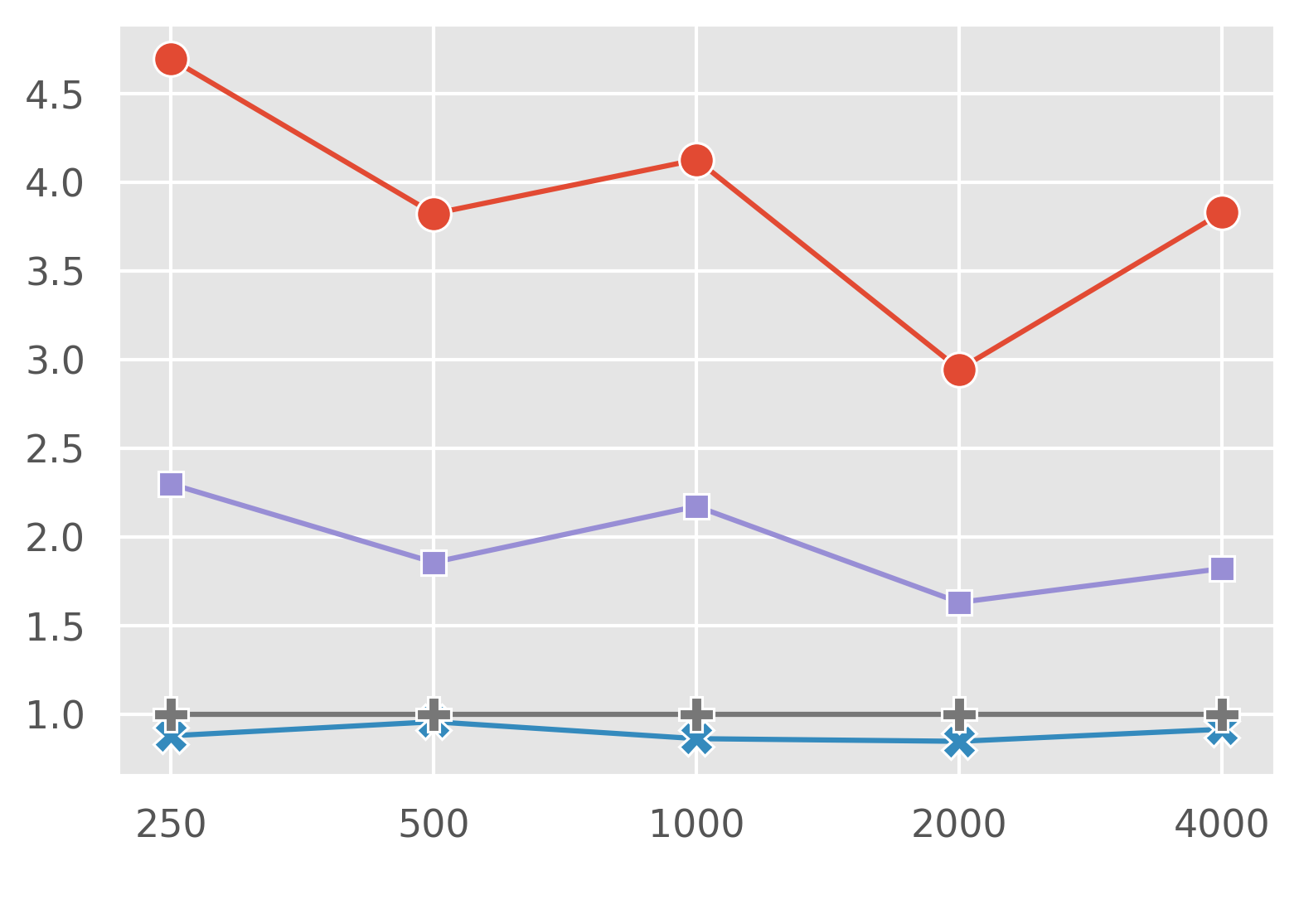}
    \end{center}
\end{minipage}
\\
\multicolumn{3}{c}{
\begin{minipage}{1.0\hsize}
\begin{center}
\caption{Estimators' performance comparison with different data size $n$}
\label{fig:data_size}
\end{center}
\end{minipage}
}
\\
\begin{minipage}{0.30\hsize}
    \begin{center}
        \includegraphics[clip, width=5.5cm]{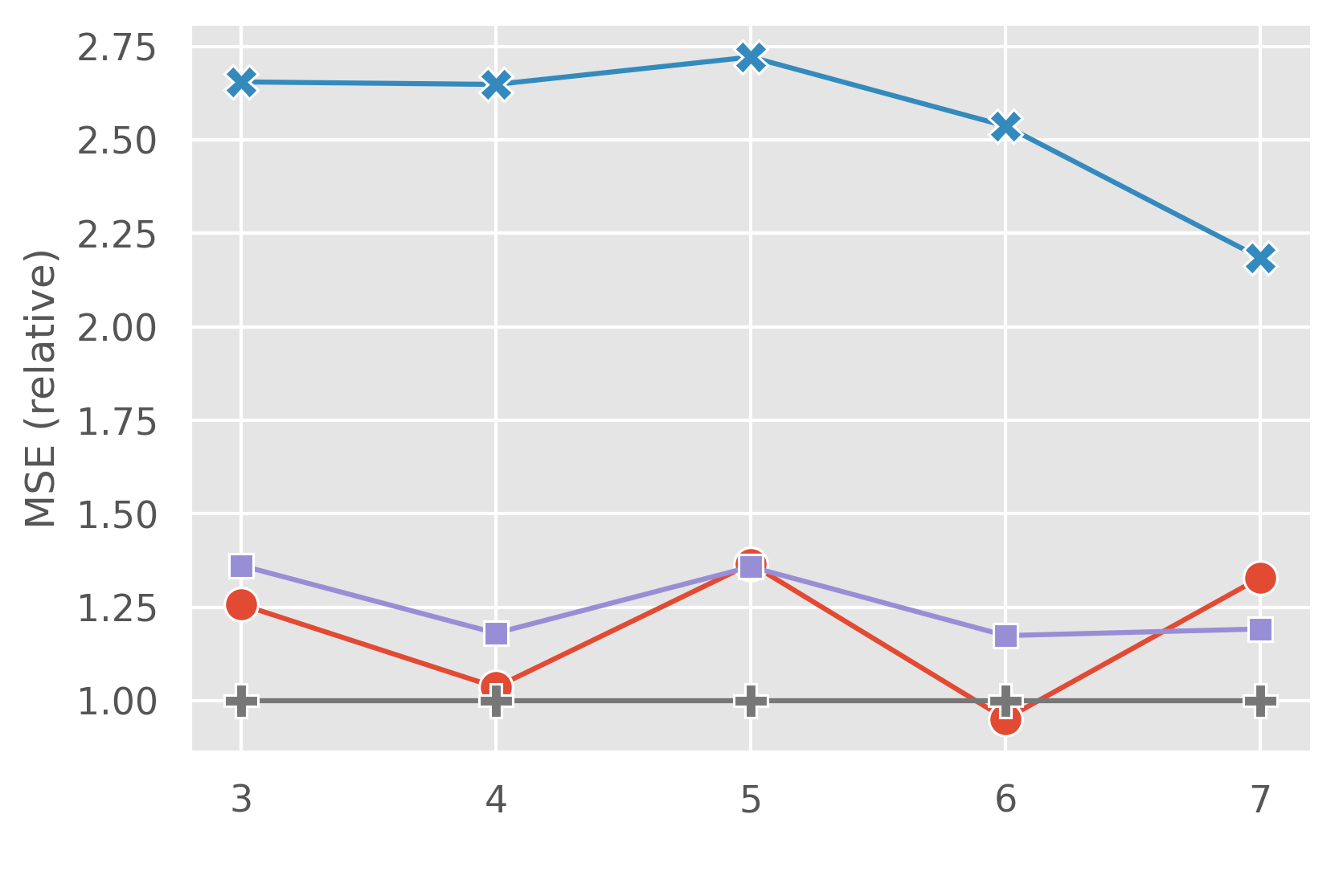}
    \end{center}
\end{minipage}
&
\begin{minipage}{0.30\hsize}
    \begin{center}
        \includegraphics[clip, width=5.5cm]{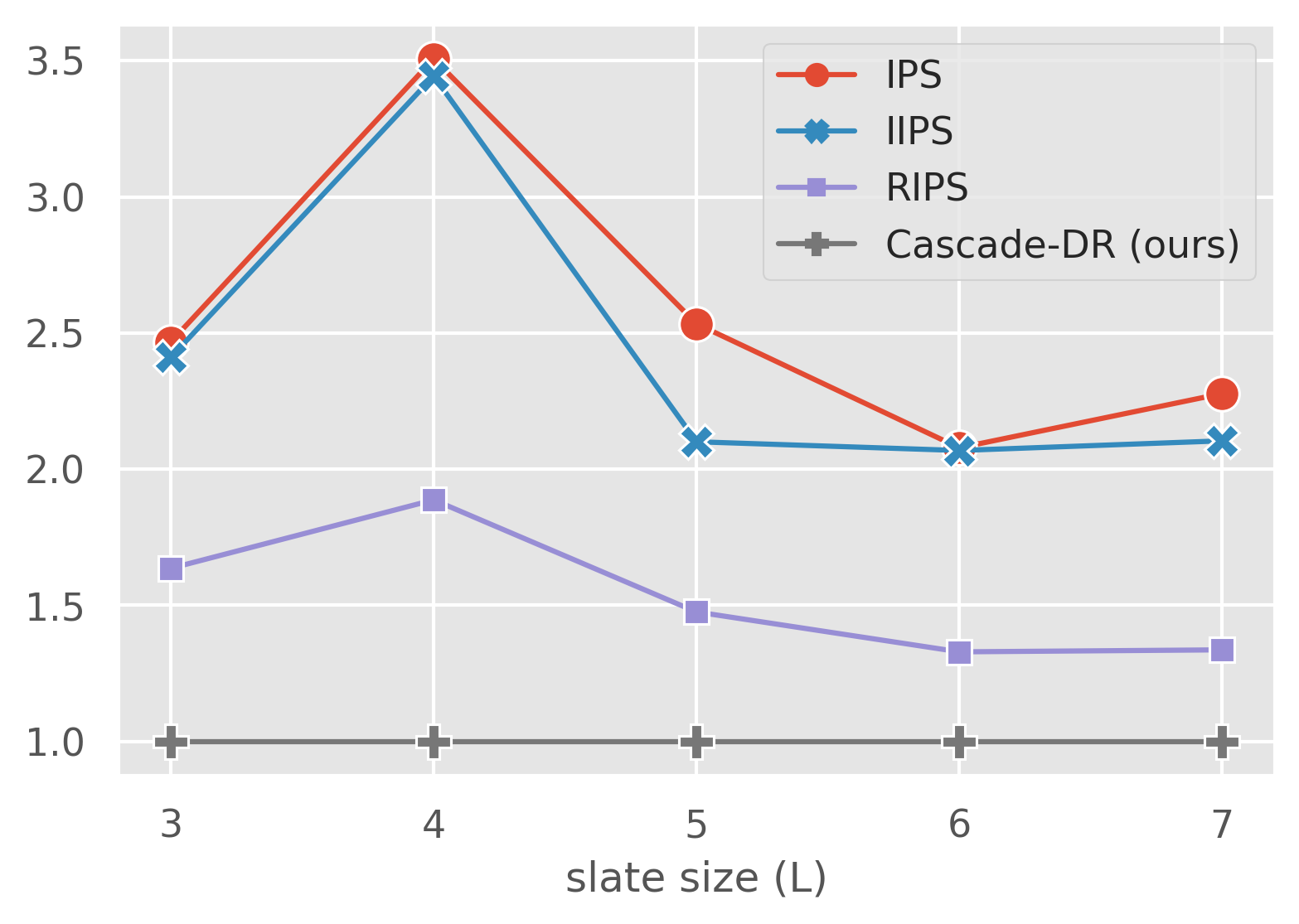}
    \end{center}
\end{minipage}
&
\begin{minipage}{0.30\hsize}
    \begin{center}
        \includegraphics[clip, width=5.5cm]{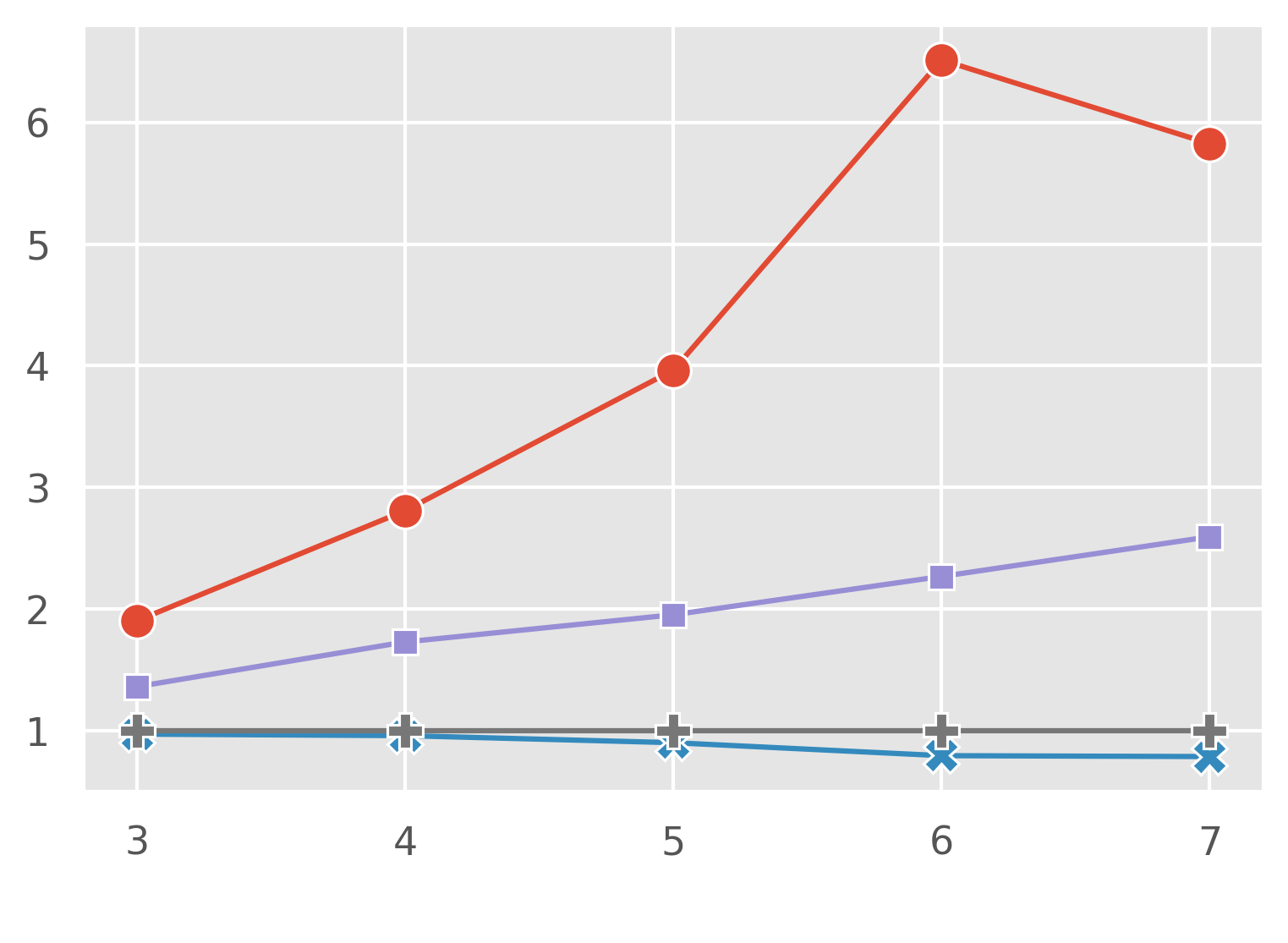}
    \end{center}
\end{minipage}
\\
\multicolumn{3}{c}{
\begin{minipage}{0.9\hsize}
\begin{center}
\caption{Estimators' performance comparison with different slate size $L$}
\label{fig:slate_size}
\end{center}
\end{minipage}
}
\\
\begin{minipage}{0.30\hsize}
    \begin{center}
        \includegraphics[clip, width=5.5cm]{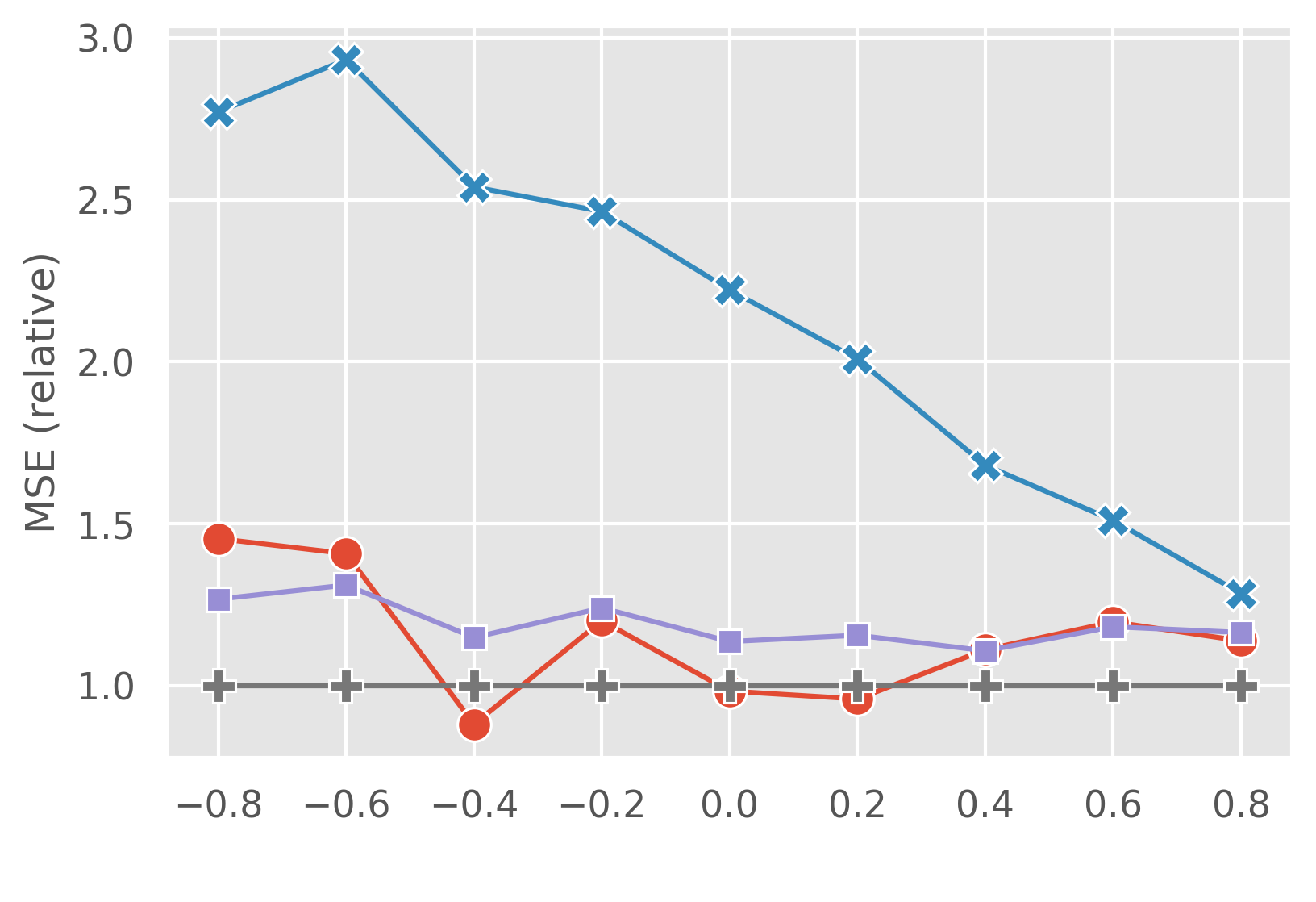}
    \end{center}
\end{minipage}
&
\begin{minipage}{0.30\hsize}
    \begin{center}
        \includegraphics[clip, width=5.5cm]{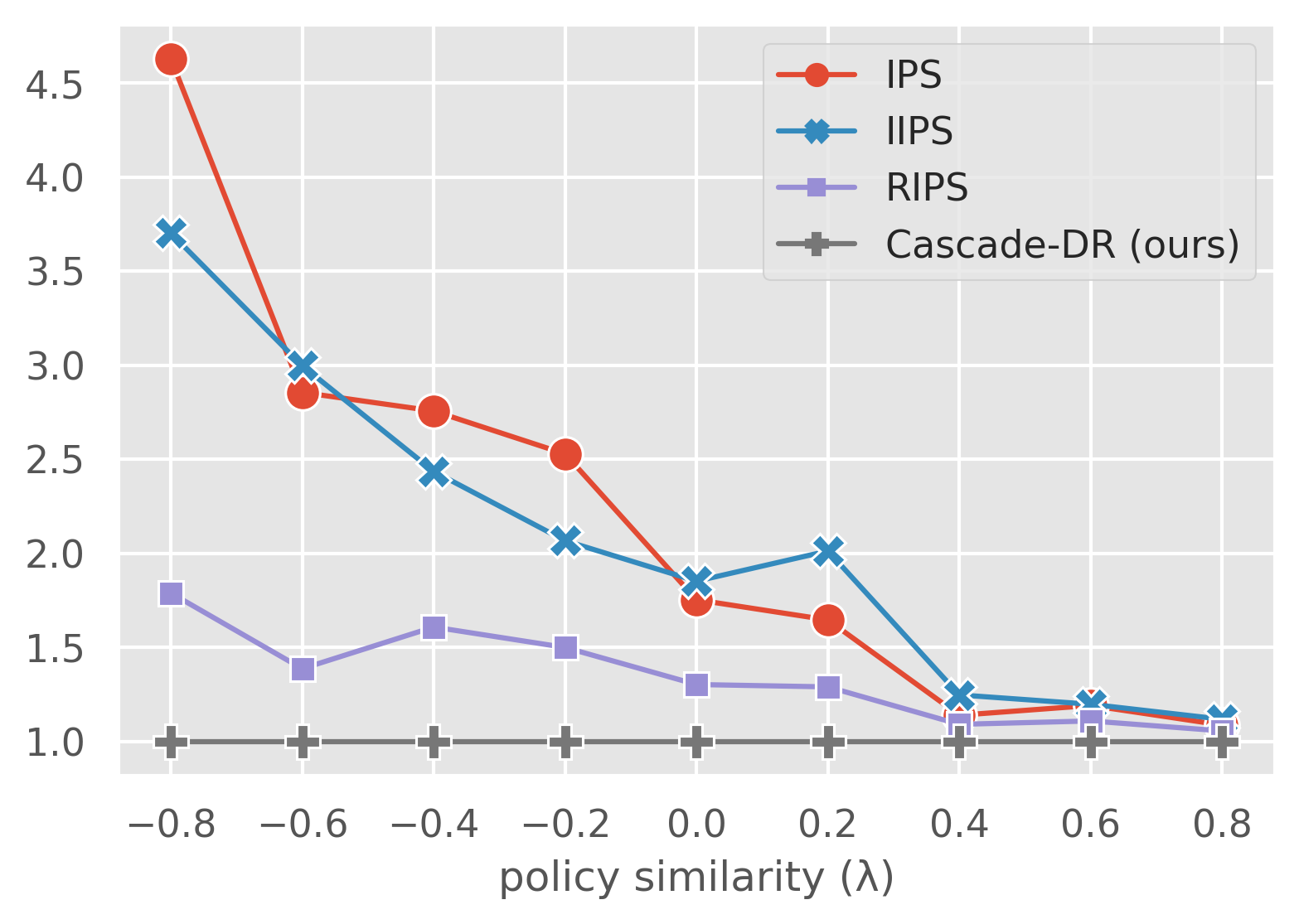}
    \end{center}
\end{minipage}
&
\begin{minipage}{0.30\hsize}
    \begin{center}
        \includegraphics[clip, width=5.5cm]{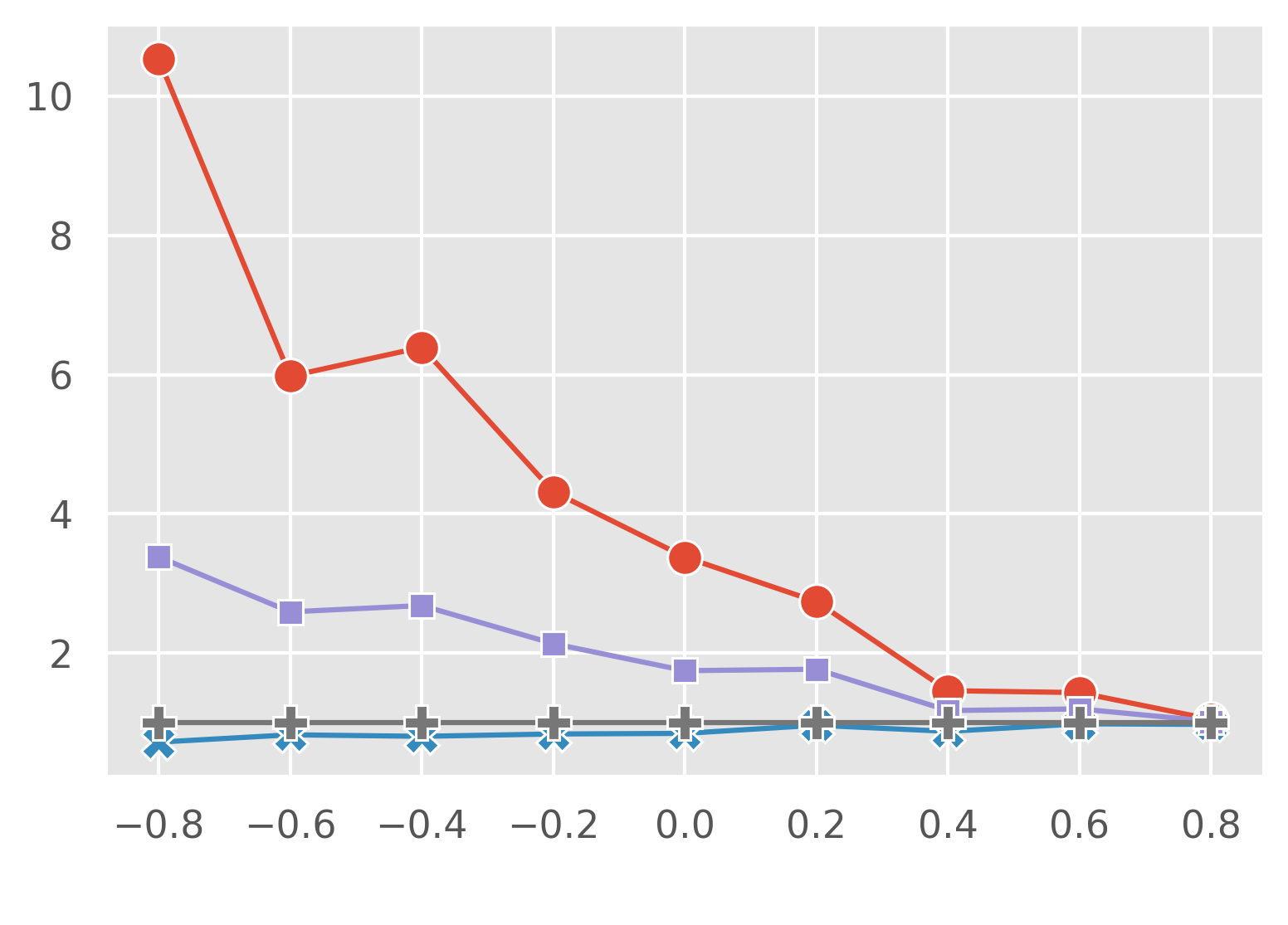}
    \end{center}
\end{minipage}
\\
\multicolumn{3}{c}{
\begin{minipage}{0.9\hsize}
\begin{center}
\caption{Estimators' performance comparison with different policy similarity $\lambda$}
\label{fig:eval_policy_sim}
\end{center}
\end{minipage}
}
\\ \bottomrule
\end{tabular}
}
\vskip 0.1in
\raggedright
\fontsize{8.5pt}{8.5pt}\selectfont \textit{Note}:
The plots compare the MSE of the estimators relative to that of Cascade-DR (i.e., $\mathrm{MSE} (\hat{V}) / \mathrm{MSE} (\DR)$).
A lower (relative) MSE indicates that the estimator is more accurate.
\end{figure*}

\subsection{Results}
\label{sec:synthetic_result}
Figures~\ref{fig:data_size}-\ref{fig:eval_policy_sim} compare the MSE of the estimators relative to that of Cascade-DR (i.e., $\mathrm{MSE} (\hat{V}) / \mathrm{MSE} (\DR)$).
In the following, we discuss the results of the three experiments \textbf{(i)-(iii)}.

\paragraph{\textbf{(i) How do the estimators perform with different data sizes?}}
Figure~\ref{fig:data_size} shows how the estimators perform as the data size $n$ grows. The result indicates that Cascade-DR is the most stable, achieving the best or the second-best MSE in all reward structures and data sizes. In particular, Cascade-DR consistently outperforms RIPS, empirically verifying its variance reduction property.

Specifically, when the reward structure is \textit{standard}, the result demonstrates that the estimators' performance depends heavily on the data size. When the data size is large, the bias is the dominant term in the MSE of the estimators. In such a situation, IPS performs the best, as it is unbiased while the others are not. When the data size is smaller, however, IPS suffers from large variance and has the worst performance, as the variance plays a crucial role in a small sample setting. On the other hand, Cascade-DR performs the best on smaller data, leveraging the cascade assumption and baseline estimator to reduce the variance. In contrast, IIPS suffers from serious bias because of its strong assumption and performs poorly with any data size.
When the reward structure is \textit{cascade}, we observe a similar trend with that of \textit{standard}. The difference is that Cascade-DR and RIPS perform better than IPS even with large data sizes (e.g., $n=2000,4000$). 
This is because, in addition to IPS, both Cascade-DR and RIPS are unbiased in this setting. Therefore, the difference in their performances is attributed to the difference in their variance. 
Finally, when the reward structure is \textit{independence}, IIPS performs the best followed by Cascade-DR. This result is because all estimators are unbiased and only the variance matters in this setting. However, the independence assumption of IIPS rarely holds in real-life scenarios. Thus, we argue that the results of the \textit{standard} and \textit{cascade} cases are more practically relevant.

\paragraph{\textbf{(ii) How do the estimators perform with different slate sizes?}}
Figure~\ref{fig:slate_size} shows the estimators' performances with varying slate size $L$. 
The result indicates that Cascade-DR performs better than the others as $L$ becomes larger when the reward structure is \textit{independence}. When the reward structure is \textit{cascade}, however, we observe that the improvement given by Cascade-DR becomes less impressive with a large slate size $L$. We attribute this observation to the accuracy of the baseline estimator $\hat{Q}$. When the underlying reward structure is complex (\textit{cascade} or \textit{standard}) and the slate size is large, we observe that ensuring the accuracy of the baseline estimator is more difficult and the benefit of using Cascade-DR is limited. Nonetheless, Cascade-DR is still the best estimator in almost all settings, but the results also suggest that there is room for improvement with respect to the baseline estimator.

\paragraph{\textbf{(iii) How do the estimators perform with different evaluation policy similarities?}}
We finally validate the effect of policy similarity $\lambda$ on the estimators' performance. Figure~\ref{fig:eval_policy_sim} shows the results with varying policy similarity $\lambda$. It is clear from the figures that Cascade-DR provides a large improvement when the behavior and evaluation policies greatly diverge. When an evaluation policy differs greatly from the behavior policy (e.g., $\lambda=-0.8,-0.6$), the importance weights blow up for both IPS and RIPS. These estimators therefore suffer from inflated variances. Moreover, in \textit{standard} and \textit{cascade}, IIPS yields severe bias especially when there is a weak overlap between the behavior and evaluation policies, as its independence assumption is greatly violated. IIPS performs better than our estimator when the reward structure is \textit{independence}, however, 
the reward structure can often be more complex in reality.

\paragraph{\textbf{Summary of the experimental findings}}
In the synthetic experiments, we observe that Cascade-DR works better than the other existing estimators when the reward structure is \textit{standard} or \textit{cascade}. It is reasonable that Cascade-DR is the best when the cascade assumption holds, as it is unbiased and has desirable variance. Moreover, even when the cascade assumption does not hold (i.e., the reward structure is \textit{standard} and Cascade-DR is biased), Cascade-DR works better than IPS when the data size is small, as it balances the bias and variance well. Moreover, we observe that Cascade-DR is robust to divergence in the behavior and evaluation policies. 
IIPS is the best when the independence assumption is valid, but this assumption rarely holds in practice. These observations lead us to conclude that the proposed estimator enables a more accurate and reliable OPE of ranking policies.

\section{Experiments: Real-World Data}
In this section, we compare the estimators using real-world data.

\paragraph{\textbf{Data Collection}}
To verify the OPE performance of the proposed estimator in a real-world application, we conducted a data collection experiment on an e-commerce platform. This platform uses data-driven algorithms to optimize a ranking of modules to showcase a set of products.\footnote{Figure~\ref{fig:module} in Appendix illustrates an interface of modules (slate action) in our application.} We collected two logged bandit datasets $\calD_A$ and $\calD_B$ by randomly assigning two \textit{factorizable} policies, $\pi_A$ and $\pi_B$, to the users on the platform. Each dataset contains $n_A=2153$ and $n_B=1968$ observations. The data sizes are small, because this is an initial attempt to implement OPE for the platform. Note that, in this application, $\mx$ is a five-dimensional user context vector, $\ma$ is a list of products presented to the user (where $|\calA|=2$), $\mr$ is a binary click indicator, and the slate size is $L=6$.

\paragraph{\textbf{Experimental procedure}}
We use $\calD_A$ to estimate the performance of $\pi_B$. This means that, in this real-world experiment, $\pi_A$ is the behavior policy and $\pi_B$ is the evaluation policy. We first approximate the ground-truth policy value of the evaluation policy $V(\pi_B)$. We do this by \textit{on-policy estimation}, i.e., $V(\pi_B) \approx V_{\mathrm{on}}(\pi_B; \calD_B) = \mE_{n_B} [ r^{\ast} ]$. We then replicate logged bandit data ${\calD_A}^{\prime}$ by \textit{bootstrap sampling} from $\calD_A$ with 20 different random seeds. We obtain a distribution of the estimators' performance by repeating the experiment with different bootstrapped samples.
Finally, we estimate the policy value of $\pi_B$ with each estimator and calculate the squared error $\mathrm{SE}(\hat{V}) = (V_{\mathrm{on}}(\pi_B; \calD_B) - \hat{V}(\pi_B; {\calD_A}^{\prime}))^2$ as the estimator's performance measure. A lower value of SE indicates that the estimator is more accurate.

\begin{figure}[tb]
    \centering
    \includegraphics[clip, width=8.5cm]{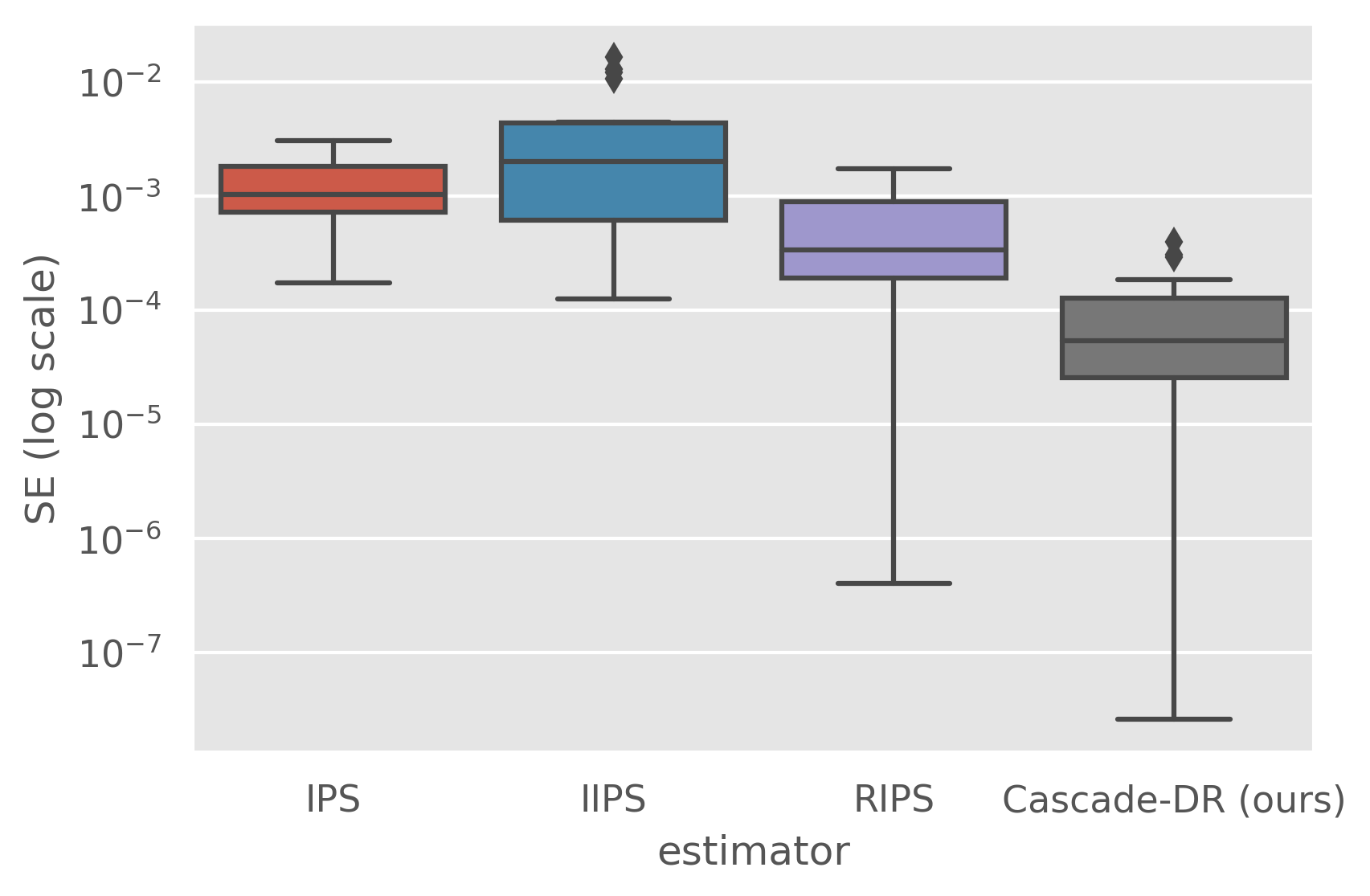}
    \caption{Box plots of the estimators' performance (log-scaled squared error) in the real-world experiment}
    \label{fig:real_result}
\end{figure}

\paragraph{\textbf{Results}}
Figure~\ref{fig:real_result} compares the estimators' SE with 20 different random seeds. The result demonstrates that Cascade-DR clearly outperforms the existing estimators, achieving lower SEs in every quartile. In addition, Cascade-DR improves the worst case performance of the other estimators by over 50\%. The results demonstrate that Cascade-DR is able to provide more accurate and stable OPE than existing estimators in the real-world application.

\section{Conclusion and Future Work}
In this paper, we studied OPE for ranking policies in the slate contextual bandit setting. 
First, we investigated the statistical properties of the existing estimators and discussed their limitations. In particular, we showed that RIPS can have a large variance, particularly when there is a weak overlap between behavior and evaluation policies and the slate size is large. To overcome the variance issue of RIPS, we propose Cascade-DR, leveraging the Markov structure of the cascade assumption. The proposed estimator reduces the variance by exploiting the baseline estimator and performing propensity weighting only on its residual. 
We showed that Cascade-DR satisfies unbiasedness under the cascade assumption. This ensures that Cascade-DR achieves a smaller bias compared to IIPS. Moreover, we proved that Cascade-DR can reduce the variance of RIPS under a reasonable assumption on the baseline estimator. Empirical results demonstrate that the proposed estimator leads to a superior OPE of ranking policies in both synthetic and real-world data.

In future work, we plan to explore methods to conduct estimator selection with logged bandit data. As we have discussed in Section~\ref{sec:synthetic_result}, an accurate estimator can change depending on the data generating process such as reward structures. Therefore, establishing a reliable data-driven method to identify an appropriate estimator will be a valuable  research direction. 

\begin{acks}
The authors would like to thank Richard Liu, Koichi Takayama, Kazuki Mogi, and Masahiro Nomura, for their helpful feedback. Additionally, we would like to thank the anonymous reviewers for their constructive reviews and discussions.
\end{acks}

\bibliographystyle{ACM-Reference-Format}
\balance{}
\bibliography{arxiv.bbl}

\clearpage
\appendix

\begin{figure}[tb]
    \centering
    \includegraphics[clip, width=5.0cm]{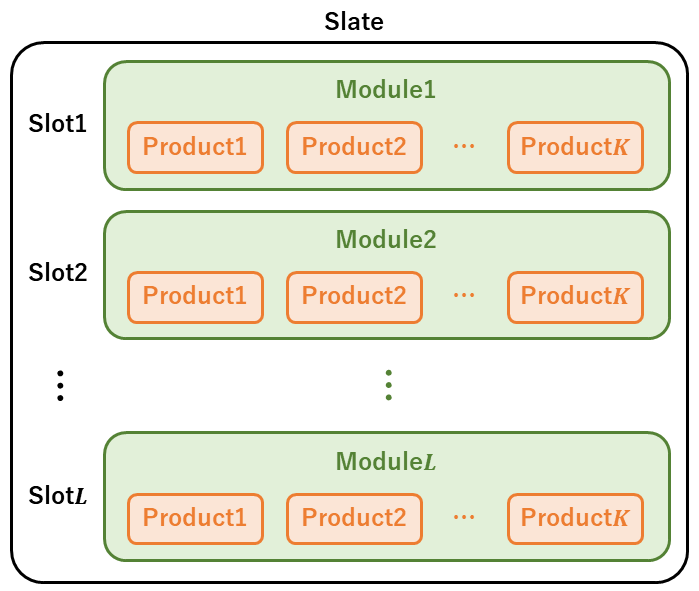}
    \caption{Modules as a slate action in the real-world data}
    \label{fig:module}
    \raggedright
    \fontsize{8.5pt}{8.5pt} \selectfont \textit{Note}: This figure illustrates a ranking of modules in the e-commerce platform used in our real-world experiment. A "Module" indicates a category of products, such as "Recommended items" or "Campaign information". A factorizable policy chooses which module to present at each slot to maximize the number of clicks observed in a ranking.
\end{figure}

\section{Omitted Proofs} \label{app:proof}

\subsection{Proof of Proposition~\ref{prop:unbiased}} \label{prof:unbiased}
\begin{proof}
To prove the unbiasedness of Cascade-DR, we first show the recursive structure of the expectation of Cascade-DR. Note that we use $\mE_{b(l)}[\cdot] := \mE_{(a_l, r_l) \sim \pi_b(a_l \mid \mx, \ma_{1:l-1})p(r_l \mid \mx, \ma_{1:l}, \mr_{1:l-1})}[\cdot \mid \mx, \ma_{1:l-1}, \mr_{1:l-1}]$ to denote the expectation over $\pi_b$ and
$\mE_{e(l)}[\cdot] := \mE_{(a_l, r_l) \sim \pi_e(a_l \mid \mx, \ma_{1:l-1})p(r_l \mid \mx, \ma_{1:l}, \mr_{1:l-1})}[\cdot \mid \mx, \ma_{1:l-1}, \mr_{1:l-1}]$ to denote that over $\pi_e$.
\begin{equation}
\begin{aligned}
    & \mE_{b(l)} \left[ \mE_{b(l+1)} \left[ \cdots \mE_{b(L)} \left[\DR^{L+1-l} \right] \right] \right] \\
    &= \mE_{b(l)} \left[ \mE_{b(l+1)} \left[ \cdots \mE_{b(L)} \left[ \frac{\pi_e(a_l \mid \mx, \ma_{1:l-1})}{\pi_b(a_l \mid \mx, \ma_{1:l-1})} \left( \alpha_l r_l + \DR^{L-l} - \hat{Q}_l \right) \right. \right. \right. \\
    & \quad \quad \quad + \mE_{a^{\prime}_l} \left[ \hat{Q}_l \right] \biggr] \biggr] \biggr] \\
    &= \mE_{b(l)} \left[ \frac{\pi_e(a_l \mid \mx, \ma_{1:l-1})}{\pi_b(a_l \mid \mx, \ma_{1:l-1})} \left( \alpha_l r_l + \left( \mE_{b(l+1)} \left[ \cdots \mE_{b(L)} \left[ \DR^{L-l} \right] \right] \right) - \hat{Q}_l \right) \right.  \\
    & \quad \quad \quad + \mE_{a^{\prime}_l} \left[ \hat{Q}_l \right] \biggr] \\
    &= \mE_{e(l)} \left[ \left( \alpha_l r_l + \left( \mE_{b(l+1)} \left[ \cdots \mE_{b(l+1)} \left[ \DR^{L-l} \right] \right] \right) - \hat{Q}_l \right) + \hat{Q}_l \right] \\
    &= \mE_{e(l)} \left[ \alpha_l r_l + \left( \mE_{b(l+1)} \left[ \cdots \mE_{b(L)} \left[ \DR^{L-l} \right] \right] \right) \right] \label{eq:unbiased_dr_recursive}
\end{aligned}
\end{equation}
Then, we derive the unbiasedness of Cascade-DR under the cascade assumption as follows.
\begin{subequations}
\begin{align}
    & \mE \left[ \dr \right] \nonumber \\
    &= \mE_{(\mx, \ma, \mr) \sim p(\mx)\pi_b(\ma \mid \mx)p(\mr \mid \mx, \ma)} \left[ \DR \right] \label{eq:unbiased_dr} \\
    &= \mE_{\mx \sim p(\mx)} \left[ \mE_{b(1)} \left[  \mE_{b(2)} \left[ \cdots \mE_{b(L)} \left[ \DR^{L+1-l} \right] \right] \right] \right] \label{eq:unbiased_dr_cascade} \\
    &= \mE_{\mx \sim p(\mx)} \left[ \mE_{e(1)} \left[ \alpha_1 r_1 + \mE_{e(2)} \bigl[ \alpha_2 r_2 + \right. \right. \nonumber \\
    & \quad \quad \quad \left. \left. \left. \cdots \mE_{e(L)} \left[ \alpha_L \mr_L + V^0 \right] \right] \right] \right] \label{eq:unbiased_dr_recursive_} \\
    &= \mE_{(\mx, \ma, \mr) \sim p(\mx)\pi_e(\ma \mid \mx)p(\mr \mid \mx, \ma)} \left[ \sum_{l=1}^L \alpha_l r_l \right] \nonumber \\
    &= V(\pi_e) \nonumber 
\end{align}
\end{subequations}
Note that from Eq.~\eqref{eq:unbiased_dr} to Eq.~\eqref{eq:unbiased_dr_cascade}, we use the recursive structure of the cascade assumption. From Eq.~\eqref{eq:unbiased_dr_cascade} to Eq.~\eqref{eq:unbiased_dr_recursive_}, we use  Eq.~\eqref{eq:unbiased_dr_recursive}. The proof also uses $V^0 = 0$.
\end{proof}

\subsection{Proof of Proposition~\ref{prop:rips} and Theorem~\ref{thrm:dr}} \label{prof:variance}
Below, we provide the proof of Theorem~\ref{thrm:dr}. Note that Proposition~\ref{prop:rips} is implied by letting $\hat{Q}=0$ in Theorem~\ref{thrm:dr}.

\begin{proof}
\begin{subequations}
\begin{align}
    & \mV_l \left( \DR^{L+1-l} \right) \nonumber \\
    &= \mE_l \left[ \left( \DR^{L+1-l} - V^{L+1-l} \right)^2 \right] \nonumber \\
    &= \mE_l \left[ \left( \DR^{L+1-l} \right)^2 \right] - \left( V^{L+1-l} \right)^2 \nonumber \\
    &= \mE_l \left[ \left( \wl \left( \alpha_l r_l + \DR^{L-l} - \hat{Q}_l \right) + \mE_{a^{\prime}_l} \left[ \hat{Q}_l \right]  \right)^2 - \left( V^{L+1-l} \right)^2 \right] \label{eq:variance_dr_a} \\
    &= \mE_l \biggl[ \Bigl( \wl \left( \alpha_l \left( r_l - q_l(\mx, \ma_{1:l}) \right) + \left( \DR^{L-l} - V^{L-l} \right)  + \left( Q_l - \hat{Q}_l \right) \right)  \nonumber \\
    & \quad \quad \quad \left. \left. + \mE_{a^{\prime}_l} \left[ \hat{Q}_l \right] \right)^2 - \left( V^{L+1-l} \right)^2 \right] \label{eq:variance_dr_b} \\
    &= \mE_l \biggl[ \Bigl( \wl \left( \alpha_l \left( r_l - q_l(\mx, \ma_{1:l}) \right) + \left( \DR^{L-l} - V^{L-l} \right) 
    \right) \nonumber \\
    & \quad \quad \quad  \left. \left. + \left( \wl \Delta_l 
    + \mE_{a^{\prime}_l} \left[ \hat{Q}_l \right] \right) \right)^2 - \left( V^{L+1-l} \right)^2 \right] \label{eq:variance_dr_c} \\
    &= \mE_l \left[ \left( \wl \left( \alpha_l \left( r_l - q_l(\mx, \ma_{1:l}) \right) \right) \right)^2 \right] \nonumber \\
    & \quad + \mE_l \left[ \left( \wl \left( \DR^{L-l} - V^{L-l} \right) \right)^2 \right] \nonumber \\
    & \quad + 2  \mE_l \left[ \wl^2 \left( \alpha_l \left( r_l - q_l(\mx, \ma_{1:l}) \right) \right) \left( \DR^{L-l} - V^{L-l} \right) \right] \label{eq:variance_dr_d} \\
    & \quad + \mE_l \left[ \left( \wl \Delta_l + \mE_{a^{\prime}_l} \left[ \hat{Q}_l \right] \right)^2 - \left( V^{L+1-l} \right)^2 \right]
    \nonumber \\
    &= \alpha_l^2 \mE_l \left[ \wl^2 \mV_{r_l} \left( r_l \right) \right]
    + \mE_l \left[ \wl^2 \mV_{l+1} \left( \DR^{L-l} \right) \right] \nonumber \\
    & \quad + 2 \alpha_l  \mE_l \left[ \wl^2 \left( r_l - q_l(\mx, \ma_{1:l}) \right) \left( \DR^{L-l} - V^{L-l} \right) \right] \nonumber \\
    & \quad + \mV_l \left( \wl \Delta_l \right) 
    \nonumber
\end{align}
\end{subequations}
Note that under the cascade assumption, $$q_l(\mx, \ma) = q_l(\mx, \ma_{1:l}) =\mE_{r_{l} \sim p(r_{l} \mid \mx,\ma_{1:l}, \mr_{1:l-1})} [r_{l} \mid \mx, \ma_{1:l}, \mr_{1:l-1} ],$$ which is independent of $\ma_{l+1:L}$.
From Eq.~\eqref{eq:variance_dr_a} to Eq.~\eqref{eq:variance_dr_b}, we use $Q_l = \alpha_l q_l(\mx, \ma_{1:l}) + V^{L-l}$. From Eq.~\eqref{eq:variance_dr_c} to Eq.~\eqref{eq:variance_dr_d}, we use conditional independence between $\wl ( \alpha_l ( r_l - q_l(\mx, \ma_{1:l}) ) + ( \DR^{L-l} - V^{L-l} ) )$ and $\wl \Delta_l + \mE_{a^{\prime}_l} [ \hat{Q}_l ]$ given $a_l$.
\end{proof}

\section{Related Work}
OPE is widely studied in recommender systems, as it enables performance estimation of new policies using only logged data, without any risky online interaction~\citep{beygelzimer2009offset, gilotte2018offline, precup2000eligibility, dudik2014doubly, jiang2016doubly, thomas2016data, farajtabar2018more, su2020doubly}.
Direct Method (DM)~\cite{beygelzimer2009offset}, IPS~\cite{precup2000eligibility, strehl2010learning}, and DR~\cite{dudik2014doubly, gilotte2018offline, jiang2016doubly, thomas2016data} are the three prevalent estimators. 
DM adopts model-based approach for the performance estimation. Specifically, DM first uses machine learning algorithms to regress the reward. Then, DM takes expectation of the estimated values over the evaluation policy. Though DM is reasonable when the reward estimator is accurate, it is prone to have serious bias because of model mis-specification~\cite{dudik2014doubly, jiang2016doubly}. In contrast, IPS uses the reward observations in the logged bandit data via importance sampling. By addressing the distributional shift between behavior and evaluation policies, IPS provides an unbiased estimation. However, IPS suffers from large variance, especially when there is a weak overlap between behavior and evaluation policies~\cite{dudik2014doubly, jiang2016doubly}. 
DR addresses the above bias-variance tradeoff by leveraging DM as a baseline and performs importance weighting only on the residual of the reward estimation. Consequently, DR achieves lower variance compared to IPS, while remaining unbiased.

Although these estimators work for single item recommendation policies, in practice, we often want to evaluate ranking policies which present a ranked list of items to users. OPE for such ranking policies remains unexplored compared to the standard contextual bandit setting. In the slate contextual bandit setting, naive applications of the prevalent OPE methods confront challenges because of a large combinatorial item space. In particular, IPS struggles with extremely large variance~\citep{swaminathan2017off, li2018offline, mcinerney2020counterfactual}. 
To tackle the variance issue, IIPS~\citep{li2018offline} and RIPS~\citep{mcinerney2020counterfactual} utilize user behavior assumptions to make combinatorial item space tractable. 
IIPS estimates the policy value under the independence assumption, which assumes that a user interacts with items independently. Under the independence assumption, the reward observed at each position
is totally independent of all the other items in the same slate. The benefit of IIPS is that it dramatically reduces the variance because of the strong independence assumption. Although IIPS is unbiased under the independence assumption, in real world data where this assumption generally does not hold, it suffers from serious bias~\cite{mcinerney2020counterfactual}.
In contrast to IIPS, RIPS estimates the policy value under the cascade assumption, which assumes that a user interacts with items sequentially from the top position to the bottom~\citep{guo2009efficient}. Under the cascade assumption, the reward for each position is dependent only on the items presented at previous positions, but is independent of the latter items. Since the cascade assumption is more realistic compared to the independence assumption, RIPS is unbiased in more cases. However, as we described in Section~\ref{sec:setup}, RIPS can still suffer from large variance, particularly when the slate size is large. 

To address the above bias-variance tradeoff between IIPS and RIPS, we proposed Cascade-DR, a DR estimator for ranking policies that works under the cascade assumption. We derived our proposed estimator inspired by DR in RL~\cite{jiang2016doubly, thomas2016data}, leveraging the structural similarities between the cascade assumption and MDP as shown in Figure~\ref{fig:assumptions}. As a result, the proposed estimator achieves better bias and variance compared to the existing estimators in the slate contextual bandit setting.

PI~\citep{swaminathan2017off, vlassis2021off} is another estimator for OPE in the slate contextual bandit setting. This estimator is designed for the situations where we can access to only the slate-level reward, $\mr^{\ast}$. Then, PI estimates the policy value as follows:
\begin{align*} \label{eq:linearity_assumption}
    \PI: = \mE_{n} [ \mE_{\pi_e} [\mathbf{1}_{\ma} \mid \mx ]^{\top} \mathbb{E}_{\pi_b} [\mathbf{1}_{\ma} \mathbf{1}_{\ma}^{\top} \mid \mx ]^{\dagger} \mathbf{1}_{\ma} r^{\ast}]],
\end{align*}
where superscript $\dagger$ indicates the pseudoinverse of the matrix.

PI is not suitable when slot-level rewards are observable, as it cannot use any information about slot-level rewards.
Furthermore, although PI is unbiased under the independence assumption, the independence assumption is usually unrealistic~\citep{mcinerney2020counterfactual}. Therefore, if the assumption does not hold, it can still lead to serious bias, as empirically verified in~\citet{mcinerney2020counterfactual}.

\section{Implementation} \label{sec:implementation}
One possible limitation of Cascade-DR is that it involves a more complicated implementation compared to the previous IPS estimators (i.e., IPS, IIPS, and RIPS). In particular, we need to implement a recursive style baseline estimation process to obtain $\hat{Q}$. To reduce this potential overhead when using Cascade-DR, we have prepared an easy-to-use implementation of our estimator in \textit{OpenBanditPipeline} (OBP)\footnote{\href{https://github.com/st-tech/zr-obp}{\texorpdfstring{\hrefcolor{https://github.com/st-tech/zr-obp}}}}. Our implementation is accessible by \textbf{obp.ope.SlateCascadeDoublyRobust}. Our public implementation allows researchers and practitioners to use our approach easily for their own purposes.

\end{document}